\documentclass[11pt]{article}

\usepackage{microtype}
\usepackage{graphicx}
\usepackage{subcaption}
\usepackage{booktabs}
\usepackage{microsoft-tech-report}
\usepackage{hyperref}
\usepackage{amsmath}
\usepackage{amssymb}
\usepackage{amsfonts}
\usepackage{mathtools}
\usepackage{amsthm}
\usepackage{bm}
\usepackage{enumitem}
\usepackage[capitalize,noabbrev]{cleveref}
\usepackage{xspace}
\usepackage{titletoc}
\usepackage{placeins}
\usepackage{tcolorbox}
\usepackage[ruled,linesnumbered,noend,noline]{algorithm2e}
\DontPrintSemicolon
\usepackage{listings}
\tcbuselibrary{theorems,skins,breakable,listings}
\usepackage{xcolor}
\usepackage{colortbl}
\usepackage{multirow}
\usepackage{float}
\usepackage{wrapfig}
\usepackage{pifont}
\usepackage{arydshln}
\usepackage[font=small,labelfont=bf]{caption}
\usepackage{nicefrac}
\usepackage{natbib}

\newcommand{\cmark}{\ding{51}}

\newcommand{\redeck}{\textsc{ReDeck}}

\newcommand{\deckquiz}{\textsc{DeckQuiz}}

\definecolor{ours}{HTML}{ADD8E6}

\definecolor{promptbg}{HTML}{F7F7F8}
\definecolor{promptborder}{HTML}{5B5B5B}
\newtcolorbox{promptbox}[1][]{
  colback=promptbg, colframe=promptborder, fonttitle=\bfseries\sffamily,
  breakable, enhanced, arc=3pt, boxrule=0.8pt,
  left=10pt, right=10pt, top=8pt, bottom=8pt,
  title={#1}
}

\lstdefinestyle{json}{
  basicstyle=\small\ttfamily,
  breaklines=true,
  breakatwhitespace=false,
  showstringspaces=false,
  columns=fullflexible,
  literate={"}{\textquotedbl}1,
}

\newif\ifappendixtoc
\appendixtoctrue

\definecolor{color_blue}{HTML}{E7EFFA}
\definecolor{color_green}{HTML}{E6F8E0}
\definecolor{color_gray}{HTML}{ECECEC}
\definecolor{pearDark}{HTML}{2980B9}
\definecolor{theoremblue}{HTML}{EBF5FB}
\definecolor{theoremborder}{HTML}{2980B9}
\definecolor{propgreen}{HTML}{EAFAF1}
\definecolor{propborder}{HTML}{27AE60}
\definecolor{defyellow}{HTML}{FEF9E7}
\definecolor{defborder}{HTML}{F39C12}
\definecolor{remarkgray}{HTML}{F2F3F4}
\definecolor{remarkborder}{HTML}{7F8C8D}

\hypersetup{
  colorlinks=true,
  linkcolor=msftblue,
  citecolor=msftblue,
  urlcolor=msftblue,
  pdftitle={ReDeck: Step-Level Render-Grounded Refinement for Document-to-Slide Generation}
}

\newtcbtheorem[auto counter]{theorem}{Theorem}{
  colback=theoremblue, colframe=theoremborder, fonttitle=\bfseries,
  breakable, enhanced, sharp corners, boxrule=0.6pt, left=6pt, right=6pt, top=6pt, bottom=6pt,
  crefname={Theorem}{Theorems}
}{thm}

\newtcbtheorem[use counter from=theorem]{proposition}{Proposition}{
  colback=propgreen, colframe=propborder, fonttitle=\bfseries,
  breakable, enhanced, sharp corners, boxrule=0.6pt, left=6pt, right=6pt, top=6pt, bottom=6pt,
  crefname={Proposition}{Propositions}
}{prop}

\newtcbtheorem[use counter from=theorem]{lemma}{Lemma}{
  colback=theoremblue, colframe=theoremborder, fonttitle=\bfseries,
  breakable, enhanced, sharp corners, boxrule=0.6pt, left=6pt, right=6pt, top=6pt, bottom=6pt,
  crefname={Lemma}{Lemmas}
}{lem}

\newtcbtheorem[use counter from=theorem]{corollary}{Corollary}{
  colback=propgreen, colframe=propborder, fonttitle=\bfseries,
  breakable, enhanced, sharp corners, boxrule=0.6pt, left=6pt, right=6pt, top=6pt, bottom=6pt,
  crefname={Corollary}{Corollaries}
}{cor}

\newtcbtheorem[use counter from=theorem]{definition}{Definition}{
  colback=defyellow, colframe=defborder, fonttitle=\bfseries,
  breakable, enhanced, sharp corners, boxrule=0.6pt, left=6pt, right=6pt, top=6pt, bottom=6pt,
  crefname={Definition}{Definitions}
}{def}

\newtcbtheorem[use counter from=theorem]{assumption}{Assumption}{
  colback=defyellow, colframe=defborder, fonttitle=\bfseries,
  breakable, enhanced, sharp corners, boxrule=0.6pt, left=6pt, right=6pt, top=6pt, bottom=6pt,
  crefname={Assumption}{Assumptions}
}{asm}

\newtcbtheorem[use counter from=theorem]{remark}{Remark}{
  colback=remarkgray, colframe=remarkborder, fonttitle=\bfseries,
  breakable, enhanced, sharp corners, boxrule=0.6pt, left=6pt, right=6pt, top=6pt, bottom=6pt,
  crefname={Remark}{Remarks}
}{rem}

\crefname{tcb@cnt@theorem}{Theorem}{Theorems}
\crefname{tcb@cnt@proposition}{Proposition}{Propositions}
\crefname{tcb@cnt@lemma}{Lemma}{Lemmas}
\crefname{tcb@cnt@corollary}{Corollary}{Corollaries}
\crefname{tcb@cnt@definition}{Definition}{Definitions}
\crefname{tcb@cnt@assumption}{Assumption}{Assumptions}
\crefname{tcb@cnt@remark}{Remark}{Remarks}
\crefname{algocf}{Algorithm}{Algorithms}

\techreportshorttitle{ReDeck}

\begin{document}
\thispagestyle{empty}

\noindent
\begin{minipage}[c]{0.5\linewidth}
\raggedright
\raisebox{-0.5\height}{\msftbrandmark}
\end{minipage}
\begin{minipage}[c]{0.49\linewidth}
\raggedleft
{\msftdatefont\small\color{msftgray}August 2026}
\end{minipage}\par
\vspace{0.35em}
\noindent{\color{msftline}\rule{\linewidth}{0.8pt}\par}

\vspace{1.0em}
\begin{center}
{{\msfttitlefont\fontsize{17}{20.5}\selectfont\color{msftdark}
\redeck: Step-Level Render-Grounded Refinement for Document-to-Slide Generation\par}}
\vspace{1.25em}

{\normalsize\rmfamily\color{msftdark}
Muzhao Tian$^{4,*\dagger}$ \hspace{0.5em}
Zezi Zeng$^{3,*}$ \hspace{0.5em}
Yifan Yang$^{1,\ddagger}$ \hspace{0.5em}
Xin Gao$^{4}$ \hspace{0.5em}
Yan Li$^{2}$ \hspace{0.5em}
Zisu Huang$^{4}$\\[-0.1em]
Xiaohua Wang$^{4}$ \hspace{0.5em}
Changze Lv$^{4}$ \hspace{0.5em}
Mingxi Cheng$^{1}$ \hspace{0.5em}
Bei Liu$^{1}$ \hspace{0.5em}
Kai Qiu$^{1}$ \hspace{0.5em}
Qi Dai$^{1}$\\[-0.1em]
Dong Chen$^{1}$ \hspace{0.5em}
Yue Dong$^{1}$ \hspace{0.5em}
Xiaoqing Zheng$^{4,\ddagger}$ \hspace{0.5em}
Ji Li$^{1}$ \hspace{0.5em}
Chong Luo$^{1}$\par
}
\vspace{0.22cm}

{\footnotesize\rmfamily\color{msftgray}
$^{1}$ Microsoft Corporation \quad
$^{2}$ Shanghai Jiao Tong University \quad
$^{3}$ Xi'an Jiaotong University \quad
$^{4}$ Fudan University\par
}
\end{center}

\vspace{0.45em}
\begin{msfttitlebox}
\setlength{\parindent}{0cm}
\setlength{\parskip}{0.14cm}
\raggedright
\nohyphens

Document-to-slide generation is challenging because slides are dense editable artifacts that require both faithful content selection and precise spatial layout.
Recent slide agents adopt iterative reflection, but typically follow a monolithic ``one version, one feedback'' loop: a slide or deck is rewritten, rendered afterward, and critiqued only at the turn boundary.
This delayed feedback makes local failures such as overflow, overlap, clipping, and off-canvas placement difficult to attribute and repair.
We propose \textbf{\redeck{}}, a step-level render-grounded refinement framework that decomposes slide revision into atomic edit actions and returns renderer-derived observations after each step, turning refinement into ``one edit, one observation.''
To balance local repair with global quality, \redeck{} uses multi-granular feedback: step-level render feedback for spatial errors, a turn-level adaptive critic for semantic and design guidance, and a submission-level gate for hard layout validation.
We further introduce \deckquiz{}, a benchmark that decouples content fidelity, spatial correctness, and design quality.
Across GPT-5.4, Claude-4.6, and Gemini-3.1, \redeck{} consistently outperforms existing slide-generation agents, and ablations confirm that feedback timing and granularity are critical for reliable slide refinement.

\vspace{0.14cm}
{\setlength{\parskip}{0.06cm}\small
{\msftmetalabel{Correspondence}\href{mailto:yifanyang@microsoft.com}{yifanyang@microsoft.com},
\href{mailto:zhengxq@fudan.edu.cn}{zhengxq@fudan.edu.cn}\par}
{\msftmetalabel{Code}\href{https://github.com/microsoft/ReDeck}{https://github.com/microsoft/ReDeck}\par}
}
\vspace{0.08cm}
{\footnotesize\rmfamily\itshape\color{msftgray}
$^*$ Equal contribution. \quad
$^{\dagger}$ Work done during an internship at Microsoft. \quad
$^{\ddagger}$ Corresponding authors.\par
}
\end{msfttitlebox}
\suppressfloats[t]

\section{Introduction}
\label{sec:intro}

Presentation slides are a central medium for communicating complex documents, including scientific papers, technical reports, and business analyses. 
Unlike plain text summaries, slides are dense visual artifacts: they must preserve source semantics, organize information into a coherent narrative, and arrange text, figures, charts, icons, and visual emphasis within a bounded two-dimensional canvas. 
Document-to-slide generation is therefore a coupled semantic-spatial problem, where content fidelity, layout validity, and visual readability must be optimized jointly.

Recent document-to-slide systems have largely followed a generate-then-render paradigm~\citep{doc2ppt2022, pptagent2025, autopresent2025, slidegen2025, deeppresenter2026}. 
Early methods attempt to produce a complete deck in one pass, for example by extracting an outline and rendering slides with predefined layouts~\citep{slidegen2025}. 
Since one-shot generation often fails to satisfy the many constraints of a polished deck, recent systems naturally introduce reflection: the agent renders a candidate deck, receives feedback from a critic, and revises the deck in the next turn, following the broader practice of LLM self-correction~\citep{cot2022, react2023, selfrefine, reflexion, webarena}. 
However, this standard refinement loop is still \emph{monolithic}: the agent commits a large slide- or deck-level rewrite, observes the rendered result only afterward, and then receives a holistic critique. 
In other words, existing agents mostly follow a ``one version, one feedback'' pattern.

\begin{figure}[t]
  \centering
  \includegraphics[width=\linewidth]{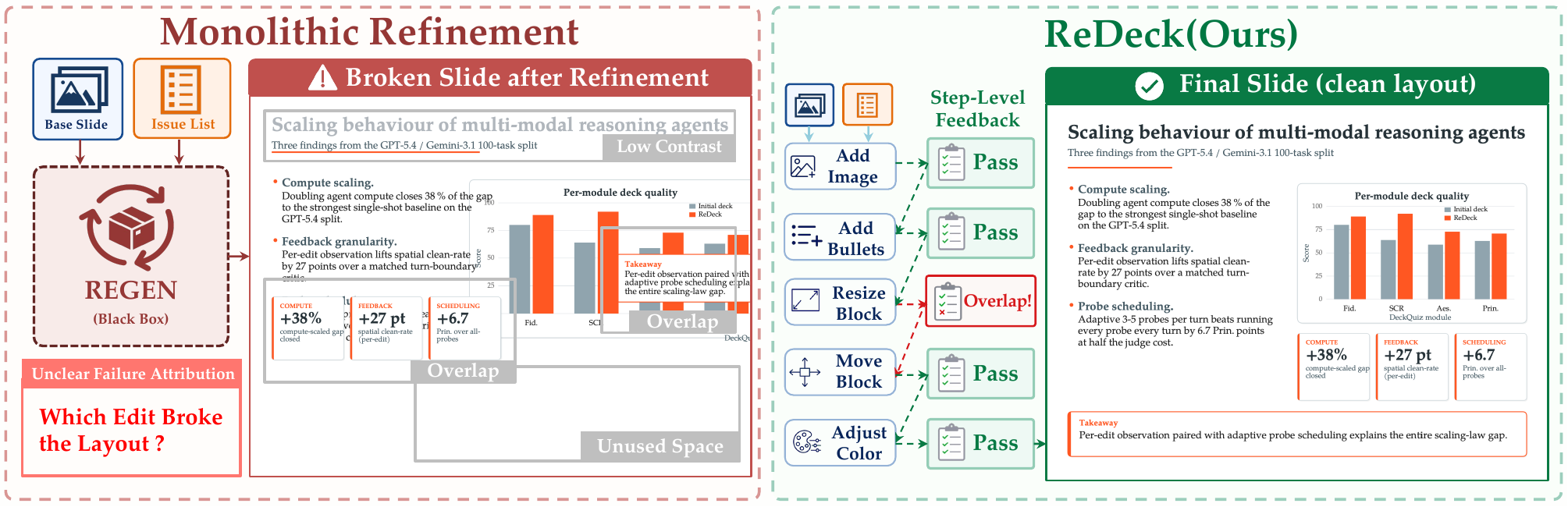}
\caption{\textbf{Turn-level reflection vs.\ \redeck{}'s step-level render-grounded refinement.}
\textit{Left}: conventional monolithic refinement commits large slide rewrites and receives feedback only after rendering the result, making overflow, overlap, low contrast, and off-canvas failures difficult to attribute to individual edits.
\textit{Right}: \redeck{} decomposes refinement into atomic edits and returns renderer-derived observations after each step, so the agent can fix local layout errors as soon as they appear while receiving global turn-level guidance.}
  \label{fig:teaser}
\end{figure}

This pattern is poorly matched to slide refinement. 
Unlike natural images, slides are editable artifacts with many small, information-bearing elements. 
A single flawed slide may contain high-level issues such as missing contributions, incorrect claims, weak narrative flow, or poor source grounding, while also containing low-level issues such as text overflow, element overlap, low contrast, unused space, clipping, and misalignment. 
Precisely describing all these issues in one global critique requires difficult alignment between visual regions, source content, and editable code. 
Even when the critique is correct, repairing many entangled issues at once is hard: the agent must decide which defect to fix first, infer which edit caused each failure, and avoid breaking already-correct regions. 
As illustrated in Fig.~\ref{fig:teaser} left, after a monolithic rewrite, the deck may contain multiple spatial failures, but their causes are no longer attributable to individual edits.

This motivates a different refinement principle: feedback should be delivered at the granularity where the corresponding problem can be most reliably observed and fixed. 
For local layout errors, waiting until the end of a turn is unnecessarily delayed. 
If adding bullets immediately creates overflow, resizing a block immediately causes overlap, or moving an image immediately pushes content off the canvas, the agent should observe that rendered consequence before taking the next action. 
Such step-level feedback reduces the critic's burden because each observation only needs to explain the consequence of the latest edit, and it reduces the editor's burden because the agent can fix one problem while the cause is still clear. 
At the same time, step-level layout feedback alone cannot judge whether the whole deck is faithful, complete, coherent, and well designed. 
Slide refinement therefore requires \emph{multi-granular feedback}: fine-grained render-grounded observations for local spatial repair, and coarser turn-level reflection for global semantic and design direction.

We introduce \redeck{}, a \emph{step-level render-grounded refinement} framework for document-to-slide generation. 
Instead of revising slides through monolithic rewrites, \redeck{} constrains the agent to a compositional action space of atomic edits, such as adding an image, inserting bullets, resizing a block, moving an element, or adjusting visual attributes. 
After each action, the rendering environment returns structured observations about the current visual state, including overflow, overlap, clipping, off-canvas placement, contrast issues, and key element geometry. 
This changes the refinement loop from ``one version, one feedback'' to ``one edit, one observation'', enabling the agent to ground its next edit in the actual rendered consequence of the previous one, as shown in Fig.~\ref{fig:teaser} right.

To coordinate local repair with global presentation quality, \redeck{} implements multi-granular feedback with separated responsibilities. 
At the step level, deterministic render feedback reports immediate spatial facts but does not act as a global judge. 
At the turn level, an adaptive deck critic maintains a persistent issue list for higher-level concerns, including narrative flow, content completeness, factual correctness, source fidelity, and visual design. 
At the submission level, a validation gate ensures that the total hard-violation count does not increase relative to the session baseline before a turn is committed. 
This design lets each feedback source operate where it is most reliable: renderer observations handle local layout failures, the deck critic handles semantic and design direction, and the submission gate ensures that temporary invalid states are not preserved as final progress.

To evaluate this framework, we introduce \deckquiz{}, a benchmark that separately measures content fidelity, spatial correctness, and design quality. 
This decomposition allows us to attribute improvements to the corresponding refinement components rather than relying only on aggregated final-deck scores. 
Across GPT-5.4, Claude-4.6, and Gemini-3.1, \redeck{} consistently outperforms existing slide-generation agents under a matched per-task LLM-call-count cap. 
Ablation studies show that step-level render grounding substantially improves spatial correctness and stabilizes refinement, while the turn-level adaptive critic provides complementary gains in semantic and design quality. 
Shared-start controls, external-benchmark evaluation, a 45-task blinded human study, and PPTX transfer test the attribution, evaluator dependence, and format scope of the proposed refinement principle.

\textbf{Contributions.}
\textbf{\textit{1) Step-level render-grounded refinement.}}
We formulate slide refinement as a sequence of atomic edit actions, each followed by renderer-derived observations, enabling local layout failures to be detected and repaired at the moment they occur.
\textbf{\textit{2) Multi-granular feedback for slide agents.}}
We combine step-level deterministic render feedback, turn-level adaptive reflection, and submission-level validation, assigning different feedback sources to the levels where they are most reliable and useful.
\textbf{\textit{3) \deckquiz{} for component-level evaluation.}}
We introduce a benchmark that decouples content fidelity, spatial correctness, and design quality, enabling fine-grained attribution of improvements in document-to-slide generation.



\section{Related Work}
\label{sec:related}


\textbf{Automated Slide Generation.}
Early document-to-slide work focused on content selection and layout from structured inputs~\citep{doc2ppt2022,d2s2021,docpres2024}. More recent systems frame slide creation as an agentic task: \textsc{AutoPresent}~\citep{autopresent2025}, \textsc{PPTAgent}~\citep{pptagent2025}, \textsc{SlideGen}~\citep{slidegen2025}, and \textsc{Auto-Slides}~\citep{autoslides} decompose the pipeline across multiple specialized agents, while \textsc{SlideTailor}~\citep{slidetailor} and \textsc{ArcDeck}~\citep{arcdeck2026} further address personalization and narrative structure. Despite these advances, most systems evaluate and refine at a coarse turn-level granularity. \textsc{DeepPresenter}~\citep{deeppresenter2026} is the most related, as it introduces rendered inspection within the refinement loop, but it does not validate layout integrity after each individual edit, allowing spatial errors to accumulate silently within a turn. Beyond slides, code-from-screenshot work in HTML/web settings~\citep{design2code2025,websight2024} addresses a closely related symbolic-to-rendered nonlinearity but operates on single-page artefacts without a multi-turn refinement contract, and the broader LLM-driven layout-generation literature~\citep{layouttransformer2021,layoutdm2023,layoutgpt2023} predicts element placements in one shot without a render-grounded repair loop.


\textbf{Feedback and Self-Correction in LLM Agents.}
Iterative refinement through feedback is a central paradigm for improving model performance, building on the reasoning-and-acting trajectory established by chain-of-thought prompting and \textsc{ReAct}~\citep{cot2022,react2023} and exemplified by frameworks such as \textit{Self-Refine} \citep{selfrefine} and \textit{Reflexion} \citep{reflexion} which utilize verbal self-correction. Grounding feedback in external evidence has been further developed in works like \textsc{Critic} \citep{critic} and related self-correction studies \citep{huang2024selfcorrection}; execution-feedback variants ground the loop in compiler or test signals~\citep{selfdebug2024,voyager2024,agentcoder2024}, while multi-agent SOP frameworks~\citep{metagpt2024,autogen2024} and software-engineering agent benchmarks~\citep{swebench2024,openhands2025} extend the same principle to richer environments. While environments like \textsc{WebArena} \citep{webarena}, \textsc{AgentBench} \citep{agentbench}, and \textsc{SWE-agent} \citep{sweagent} demonstrate the necessity of real-time observations, and GUI / vision agents that ground actions in rendered screenshots~\citep{cogagent2024,seeact2024} make the same point in the visual modality, slide agents often lack a specific rendered-state delta to capture immediate layout regressions. Existing systems predominantly rely on holistic or scalar reflection \citep{autopresent2025,arcdeck2026,deeppresenter2026}. In contrast, \redeck{} decouples feedback into step-level objective facts and turn-level semantic guidance to stabilize long-horizon refinement.

\textbf{Benchmarks for Document Design and Layout.}
Accurate evaluation of generated presentations requires moving beyond generic text similarity \citep{pptc2023,autopresent2025,pptagent2025,slidevqa2023} toward metrics that preserve source fidelity and canvas integrity. Recent benchmarks such as \textsc{PresentBench} \citep{presentbench2026}, \textsc{SlidesGen-Bench} \citep{slidesgenbench2026}, \textsc{ArcDeck} \citep{arcdeck2026}, \textsc{AeSlides} \citep{aeslides2026}, and \textsc{DECKBench} \citep{deckbench2026}, alongside cross-format suites like \textsc{BizGenEval} \citep{bizgeneval2026}, have introduced verifiable checks for layout dimensions, whitespace, visual imbalance, and reference-aligned content fidelity. These developments align with broader research on multi-constraint instruction following \citep{recast2025} and query-specific rubrics \citep{deeprubric2026}, emphasizing that evaluation signals should be instance-specific. Rooted in classical principles of perceptual organization and multimedia learning \citep{wertheimer1938laws,mayer2009multimedia,tufte1983,lidwell2010universal}, our proposed \deckquiz{} addresses the variance and bias inherent in \textsc{LLM-as-a-judge} metrics~\citep{mtbench2023,chatbotarena2024,prometheus2_2024,benchmark2_2026,rewardhacking2026} by separating hard spatial-violation criteria from subjective design principles.

\section{Method}
\label{sec:method}

\begin{figure}[t]
  \centering
  \includegraphics[width=\linewidth]{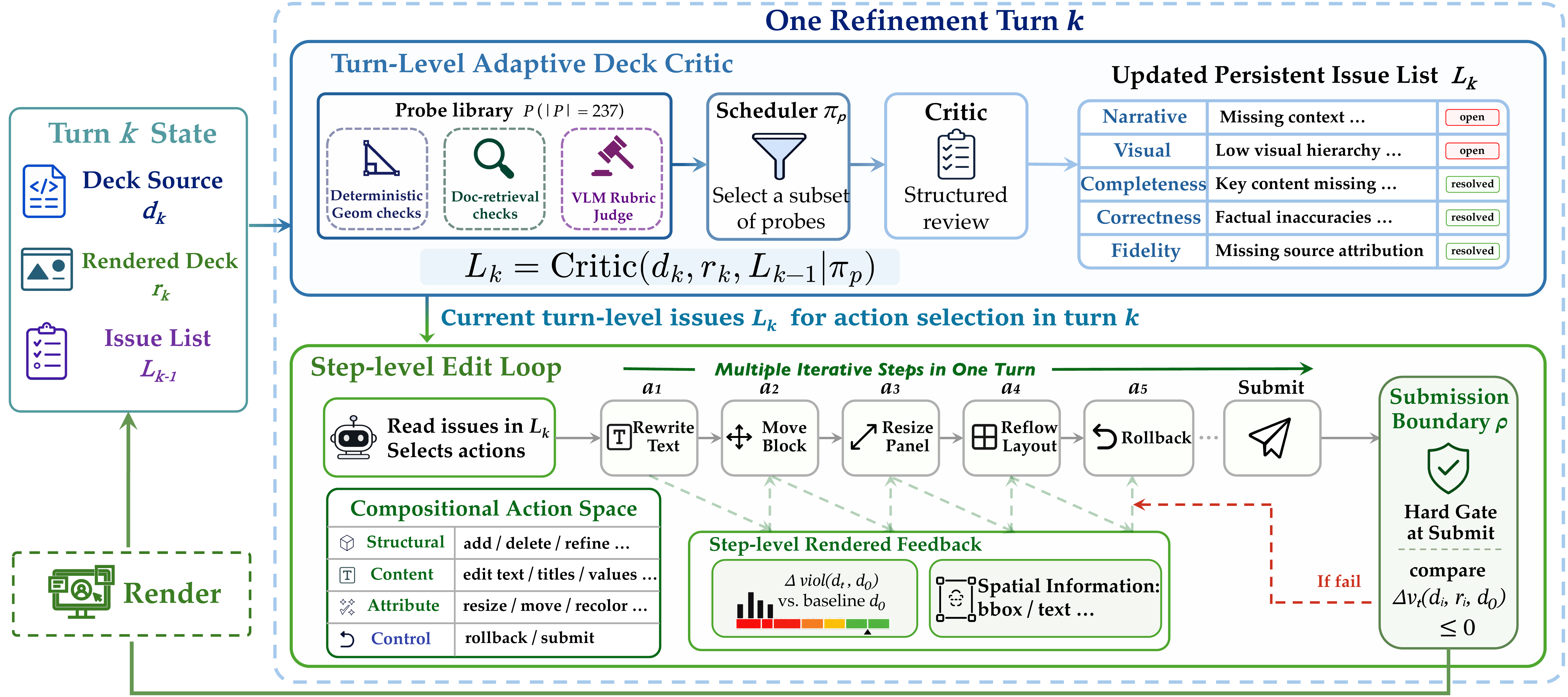}
\caption{\textbf{\redeck{} pipeline.}
  Each refinement turn nests two feedback scales over a partially observable decision process $(S, A, \Omega)$.
  \textbf{Turn lane:} Adaptive Deck Critic uses a scheduler $\pi_{\mathcal{P}}$ to draw a probe subset from library $\mathcal{P}$ and updates the persistent issue list $L_{k}$.
  \textbf{Edit lane:} after every atomic action $a_t$ from the compositional action space, the rendering environment returns render-state observation $o_t = (\Delta_{\mathrm{viol}}(d_t, d_0), A_t)$ to the next reasoning step.
  Step-level render feedback is observation, not verdict; the only hard gate is the submission boundary $\rho$, comparing the rendered violations of the submitted deck against session-start baseline $d_0$.}
  \label{fig:pipeline}
\end{figure}

In this section, we present \redeck{}, a step-level render-grounded refinement framework designed to resolve the perception-action gap in slide generation caused by highly nonlinear spatial rendering. By splitting supervision into multi-granular feedback, consisting of an inner step-level render-feedback channel and an outer turn-level critic (Figure~\ref{fig:pipeline}), \redeck{} reduces the feedback delay from $O(\text{turn})$ to $O(\text{edit})$. This dual-scale architecture efficiently grounds individual edits in rendered reality while relying on the critic for comprehensive design direction. The remainder of this section is organized as follows: we first formalize the refinement environment and define a compositional action space to prevent untraceable layout errors (\S\ref{sec:method-3.1}); we then detail the core dual-level feedback mechanisms (\S\ref{sec:method-3.2} and \ref{sec:method-3.3}); and finally, we describe the complete \redeck{} loop that systematically integrates both scales under strict validation gates (\S\ref{sec:method-3.4}).

\subsection{Problem Modeling and Compositional Action Design}
\label{sec:method-3.1}

The iterative generation of slide decks involves transitioning between a \emph{semantic reasoning space} and a \emph{rendered spatial space}. We model this process within a single turn as a sequence of discrete steps. Let $t$ denote the inner step index. At step $t$, the dynamic state of the rendering environment is represented as the tuple: $s_t = (d_t, r_t, C_t)$, where $d_t$ is the editable deck representation (e.g., HTML/CSS or structured presentation objects), $r_t$ is the spatial geometry produced by the active rendering stack (e.g., browser DOM boxes for HTML or shape bounds for PPTX), and $C_t$ contains compiler or export diagnostics (e.g., syntax errors or asset loading failures). Because the agent only edits $d_t$, it cannot directly perceive the physical layout $r_t$ or diagnostics $C_t$ without rendering. To avoid overwhelming the agent's context window with high-dimensional spatial coordinates and raw logs, the environment filters $r_t$ and $C_t$ to produce a compact, text-based observation $o_t \in \Omega$. This step-level observation $o_t$ acts as a filtered report that summarizes only the immediate technical failures and critical element coordinates resulting from the latest edit.

At the start of each refinement turn, the agent is also provided with a persistent issue list $L$. This list contains high-level design or content defects (e.g., overlapping text or missing citations) and remains static throughout the inner steps. At each step $t$, the agent uses the active observation $o_t$ alongside this static issue list $L$ to select its next action $a_t \in A$.

To handle the highly nonlinear mapping from the source code $d_t$ to the rendered geometry $r_t$~\citep{design2code2025,layoutgpt2023}, \redeck{} constrains the agent to a compositional action space $A$ of atomic edits. Rather than rewriting an entire slide in one step, the agent performs discrete modifications across three orthogonal categories: \emph{structural actions} for global element arrangements (e.g., adding or removing shapes, reflowing layout), \emph{content actions} for updating semantic information (e.g., editing text, replacing figures), and \emph{attribute actions} for precise geometric adjustments (e.g., resizing, repositioning).%
\footnote{At the tool level, these logical action types are implemented through a general-purpose \texttt{apply\_edits} tool that performs targeted search-and-replace operations on the slide source code, along with \texttt{verify\_layout}, \texttt{rollback}, and \texttt{submit} control tools. The agent is prompted to perform one logical edit per tool call; the three-way taxonomy describes the \emph{intent} of each edit rather than distinct tool endpoints.}
This atomic design isolates the spatial consequences of each edit, ensuring that any layout violation reported in $o_t$ can be directly attributed to a single, identifiable action. The action set is completed by \texttt{rollback}, which allows the agent to revert unsuccessful edits and use the \texttt{edit$\to$observe$\to$revert$\to$re-attempt} loop as a deliberate local search reminiscent of tree-search reasoning~\citep{tot2023}, and \texttt{submit}, which concludes the current turn. This formulation provides the structural foundation for the dual-level feedback mechanisms described in the following sections.

\subsection{Inner Loop: Step-Level Render Feedback}\label{sec:method-3.2}

To help the agent see the direct results of its actions, \redeck{} evaluates each atomic action $a_t$ immediately, drawing the step-level supervision idea from process-reward modelling for reasoning chains~\citep{lightman2024prm,mathshepherd2024} and porting it from \emph{reasoning steps} to \emph{rendering steps}. Instead of waiting until the end of a turn, the environment compiles the deck after every single edit and returns a step-level observation $o_t \in \Omega$ to the next step. This per-step loop creates a continuous cycle of editing, rendering, and observing, which helps detect layout errors before they accumulate. The step-level observation is defined as:
{\small
\[
o_t = \bigl(\Delta_{\text{viol}}(d_t,\ d_0),\ A_t\bigr), \quad \Delta_{\text{viol}}(d_t,\ d_0) := |\mathcal{V}(d_t)| - |\mathcal{V}(d_0)|
\]
}
where the violation delta $\Delta_{\text{viol}}$ measures the change in compiler failures and DOM-level errors (such as text overflow, overlap, and clipping) compared to the initial baseline $d_0$. To help the agent locate these errors, the environment provides this delta along with specific details, including the IDs and bounding boxes of the failing elements. The layout anchor $A_t$ records the exact positions $(x, y)$, dimensions $(w \times h)$, and text previews of the main elements, providing a clear numerical grid for planning while filtering out visual noise like page numbers. We explicitly exclude screenshots and subjective design scores at this step; step-level screenshots often cause the agent to hallucinate, while subjective scores are too inconsistent to guide individual edits.

During the editing process, this step-level feedback acts as a soft hint rather than a strict constraint. This soft policy is necessary because fixing a layout often requires going through temporary error states, such as a textbox temporarily overflowing before the agent can shrink its font. However, to prevent the agent from submitting a slide with unfixed errors, \redeck{} applies a hard check at the end of the turn when the agent calls the \texttt{submit} action. This is controlled by a simple validation rule:
{\small
\[
\rho(s_t, a_t, s_{t+1}) = \begin{cases} \mathbb{1}\!\left[\Delta_{\text{viol}}(d_{t+1},\ d_0) \le 0\right] & a_t = \mathtt{submit}, \\ 1 & \text{otherwise.} \end{cases}
\]
}
Since this rule uses the same baseline $d_0$ as the step-level observations, the submission check is highly predictable. If $\rho = 0$, the submission is rejected, and control returns to the agent with the current observation, forcing it to reduce the violation count to at most the baseline level before the turn can end.

\subsection{Outer Loop: Turn-Level Adaptive Deck Critic}\label{sec:method-3.3}

While step-level feedback focuses on individual edits, the agent still needs a broad, deck-wide view of which parts of the presentation are unsatisfactory. The Adaptive Deck Critic provides this view by evaluating both the slide source code $d$ and its rendered images $r$ at the start of each turn. This process updates a persistent issue list $L$ defined as:
{\small
\[
L \leftarrow \mathrm{Critic}(d,\ r,\ L;\ \pi_{\mathcal{P}})
\]
}
where $\pi_{\mathcal{P}}$ is the scheduling policy. Rather than giving free-form advice, the list $L$ consists of structured entries. Each entry contains the issue category, the affected slide ID, a pointer to the evidence, and a status marker. The critic monitors five key categories: narrative flow, visual layout, completeness, correctness, and source fidelity. On each turn, the critic either adds new issues to $L$ or updates the status of existing ones to represent whether they are still present, resolved, or regressed. This structured approach ensures that resolving an issue requires a formal verdict from the critic rather than a simple self-report by the agent.

\textbf{Probe library.} To detect these diverse problems, the critic uses a library $\mathcal{P}$ of 237 single-purpose probes. These probes are based on published slide benchmarks, classic visual design principles, and multimedia learning guidelines. Each probe is implemented in one of three ways: geometric checks (such as finding text overlaps or font size errors), source-retrieval checks (using BM25 index lookups to verify factual claims), or rubric-based multimodal evaluations. This structure ensures that LLM critiques enter the loop as formal, typed entries rather than unstructured conversation. To maintain feedback stability, the scheduler $\pi_{\mathcal{P}}$ dynamically activates a small budget of three to five probes per turn. The scheduler ranks probes based on which slides were recently edited, previous regression history, and the need to recheck resolved issues. Running all 237 checks on every turn would create high judge variance, causing the critic to find inconsistent, minor issues that lead to conflicting edits and prevent the layout from converging.

\subsection{The Full \redeck{} Framework}
\label{sec:method-3.4}

At the initiation of each refinement turn $k$, the adaptive deck critic evaluates the current deck source $d_{k, 0}$ to update the persistent issue list $L_k$, which remains static throughout the turn. The agent then executes an inner sequence of atomic edits in a thought--action--observation trajectory~\citep{react2023}, consuming both $L_k$ and the active step-level observation $o_t$ to select each action $a_t \in A$, until it invokes the \texttt{submit} action to trigger the validation gate $\rho$. Upon a successful submission, the final compiled source $d_{k, T_k}$ is persisted as the baseline for the next turn, setting $d_{k+1, 0} = d_{k, T_k}$; a failed submission returns control to the agent for further corrective adjustments.

To prevent optimization drift, the two scales operate under strictly decoupled write permissions: the turn-level critic holds exclusive authority to append new issues, bind localization evidence, or verify the resolution of existing entries in $L_k$, whereas the agent has read-only access to these fields and the high-frequency step-level channel serves strictly as an instantaneous spatial report without modifying $L_k$. This asymmetric information barrier prevents the agent from self-reporting success, keeping $L_k$ an objective, environment-enforced target. The dual-scale synergy is essential for convergence: without the critic, the agent over-optimizes local layouts at the expense of deck-level coherence; without step-level spatial feedback, it edits the underlying source code blindly. We empirically evaluate this dual-scale topology in \S\ref{sec:experiments}.

\section{Experiments}
\label{sec:experiments}

\subsection{Setup}
\label{sec:exp-setup}

\noindent\textbf{Task and benchmarks.}
We target \emph{scientific presentation generation} from full papers, the dominant evaluation setting in prior slide-agent work~\citep{slidegen2025,slidetailor,deeppresenter2026,pptagent2025};
cross-domain transfer across the five \textsc{PresentBench} domains is in Sec.~\ref{sec:exp-stress}.
\deckquiz{} (ours) bundles four orthogonal modules:
\textbf{ContentQuiz} (\textsc{Fid.}, four QA types: contribution / method / experiment / limitation), \textbf{SpatialCheck} (\textsc{SCR}, VLM-judged clean-rate over four hard-violation families), \textbf{Aesthetics} (\textsc{Aes.}, $0$--$5$ ordinal over five design axes), and \textbf{DeckDesign} (\textsc{Des.}, $0$--$3$ Likert over five info-architecture principles in deck and slide layers; App.~\ref{app:principles}); the four modules ask, respectively, \emph{what} facts a deck conveys, \emph{whether} they render cleanly, \emph{how each slide looks}, and \emph{how the deck organises information}.
The in-domain split contains 100 papers stratified by discipline (CS/ML 25, Bio-Med 15, Physical-Sci 25, Econ/Finance 25, Soc-Sci/HCI 10) and complexity (30 short / 40 medium / 30 long); construction details, source freeze, and per-task call budget are in App.~\ref{app:deckquiz}.
\textsc{PresentBench}~\citep{presentbench2026} is reserved for the cross-domain stress test (Sec.~\ref{sec:exp-stress}, App.~\ref{app:presentbench}); excluded from the main table due to known judge cross-model drift.

\noindent\textbf{Models, comparisons, and protocol.}
We hold the agent prompt and tool budget fixed across GPT-5.4, Gemini-3.1, and Claude-4.6, and compare \redeck{} against three published deck-level slide agents rerun end-to-end under matched per-task LLM-call-count cap: \textsc{SlideGen}~\citep{slidegen2025}, \textsc{SlideTailor}~\citep{slidetailor} (code from \texttt{nusnlp/SlideTailor}), and \textsc{DeepPresenter}~\citep{deeppresenter2026} (code from \texttt{icip-cas/PPTAgent}; the most recent system, descended from \textsc{PPTAgent}~\citep{pptagent2025}).
In-house refinement strategies (\textsc{Single-Shot}, \textsc{Self-Refine}~\citep{selfrefine}, \textsc{Reflexion}~\citep{reflexion}, per-turn \textsc{Screenshot}) appear in Sec.~\ref{sec:exp-abl} as feedback-paradigm ablations sharing our parser, tool set, and prompt scaffolding so that the contrast isolates the feedback channel.
Each (model, system, task) cell uses three seeds; main-table entries are means, with seed variability and task-paired inference reported in App.~\ref{app:stats}.
All systems run under a matched $150$ LLM-call-count cap (App.~\ref{app:budget}); token volume, cost, and latency are not matched and are reported separately in App.~\ref{app:cost}.
The four \deckquiz{} VLM judges run on a single fixed GPT-5.4 backbone for all evaluated systems, ensuring consistent scoring across rows~\citep{mtbench2023}, with deterministic decoding (App.~\ref{app:judge-config}).
Component, feedback-frequency, scheduling, transfer, and human-validation analyses are reported in Sec.~\ref{sec:exp-abl}--\ref{sec:exp-stress}.

\subsection{Main results}
\label{sec:exp-main}

\begin{figure}[H]
  \centering
  \begin{minipage}{\linewidth}
    \centering
    \captionof{table}{Performance comparison with published slide agents on \deckquiz{} (3 models $\times$ 100 tasks $\times$ 3 seeds) under matched per-task LLM-call-count caps. Best results per column are in \textbf{bold}.}
    \label{tab:main}
    \small
    \begin{tabular}{l l c c c c}
      \toprule
      Model & System &
        Fid.\ $\uparrow$ & SCR $\uparrow$ & Aes.\ $\uparrow$ &
        Des.\ $\uparrow$ \\
      \midrule
      \multirow{5}{*}{GPT-5.4}
        & SlideGen              & 65.3 & 83.4 & 3.08 & 50.2 \\
        & SlideTailor           & 60.2 & 88.2 & 2.90 & 56.4 \\
        & DeepPresenter         & 76.8 & 66.8 & 3.47 & 55.1 \\
        \cdashline{2-6}
        & \redeck{} (T0, before repair) & 80.4 & 64.1 & 2.95 & 62.8 \\
        & \redeck{} (ours)      & \textbf{88.6} & \textbf{91.5} & \textbf{3.64} & \textbf{71.2} \\
      \midrule
      \multirow{5}{*}{Gemini-3.1}
        & SlideGen              & 61.5 & 80.1 & 2.92 & 47.3 \\
        & SlideTailor           & 56.8 & 84.7 & 2.71 & 53.2 \\
        & DeepPresenter         & 72.4 & 63.5 & 3.31 & 52.0 \\
        \cdashline{2-6}
        & \redeck{} (T0, before repair) & 76.8 & 61.2 & 2.81 & 59.4 \\
        & \redeck{} (ours)      & \textbf{85.1} & \textbf{88.2} & \textbf{3.48} & \textbf{68.1} \\
      \midrule
      \multirow{5}{*}{Claude-4.6}
        & SlideGen              & 63.2 & 81.6 & 3.01 & 48.6 \\
        & SlideTailor           & 58.4 & 86.2 & 2.82 & 54.7 \\
        & DeepPresenter         & 74.5 & 64.9 & 3.38 & 53.4 \\
        \cdashline{2-6}
        & \redeck{} (T0, before repair) & 78.5 & 62.5 & 2.88 & 61.0 \\
        & \redeck{} (ours)      & \textbf{86.8} & \textbf{89.6} & \textbf{3.55} & \textbf{69.5} \\
      \bottomrule
    \end{tabular}
  \end{minipage}

  \vspace{0.45cm}
  \includegraphics[width=0.94\linewidth]{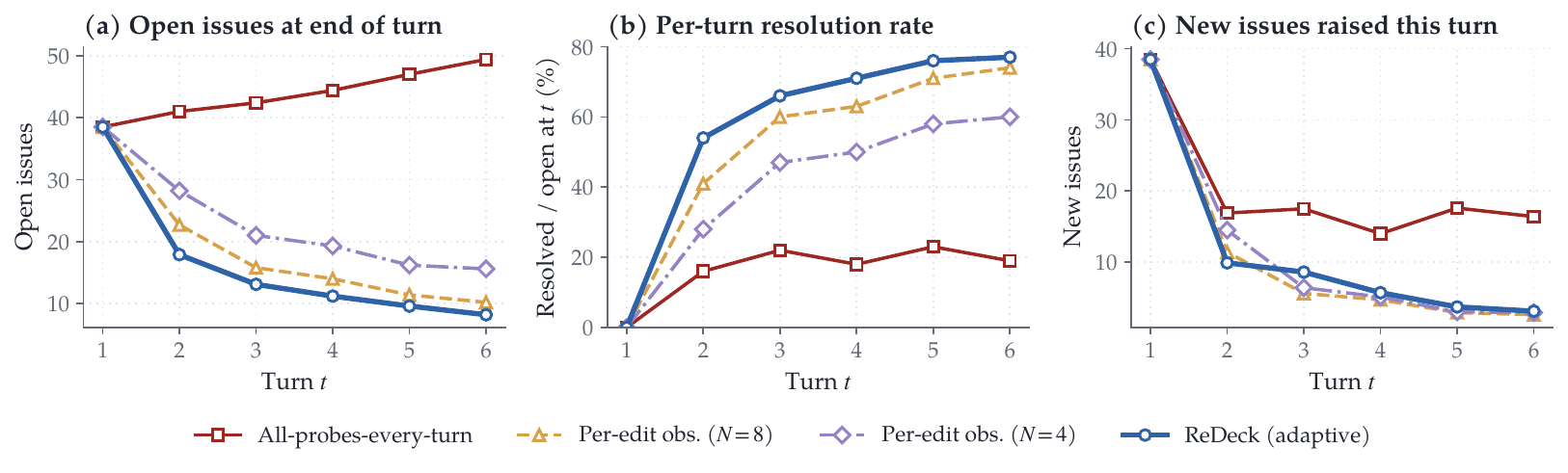}
  \caption{\textbf{Refinement convergence dynamics across scheduling and observation-frequency variants.}
    \textbf{(a)} Open-issue count per turn: \redeck{}'s adaptive scheduler drops sharply between turns~$0$--$1$ then decays to a low plateau; all-probes-every-turn drifts \emph{upward}; $N{\in}\{4,8\}$ decay but plateau higher.
    \textbf{(b)} Per-turn resolution rate: \redeck{} climbs steepest, plateauing ${\sim}77\%$; per-step variants $N{=}4{\sim}74\%$, $N{=}8{\sim}60\%$; all-probes flattens at ${\sim}20\%$.
    \textbf{(c)} New issues raised this turn: near-zero for \redeck{}, plateaus high for all-probes, and decays slowly for low-frequency variants.
    $n{=}300$ ($100 \times 3$ seeds); final quality in Tab.~\ref{tab:ablation-quality}.}
  \label{fig:dynamics}
\end{figure}

Tab.~\ref{tab:main} reports the four \deckquiz{} modules on the in-domain $100$ tasks $\times$ $3$ seeds, comparing \redeck{} against three published slide agents under the matched per-task LLM-call-count cap.
The ablations in Sec.~\ref{sec:exp-abl} attribute the gap to specific components of the loop.

Tab.~\ref{tab:main} shows that \redeck{} achieves the highest score on all four \deckquiz{} modules for each agent backbone. GPT-5.4 task-paired inference confirms all 12 comparisons after Holm correction, while the other two backbones serve as descriptive replications (App.~\ref{app:stats}).
Because \redeck{}'s initial draft is already strong on \textsc{Fid.}\ and \textsc{Des.}, we also evaluate both refinement procedures from the ReDeck and DeepPresenter initial decks. The refinement advantage persists across both starts (App.~\ref{app:stats}, Tab.~\ref{tab:attribution-inference}). Component attribution is in Tab.~\ref{tab:ablation-quality}, and a real repair trajectory is presented in App.~\ref{app:case}.

\subsection{Ablations}
\label{sec:exp-abl}
\label{sec:exp-loop-topology}

All ablations use GPT-5.4 on the same 100 tasks and three seeds as the corresponding main result.
Tab.~\ref{tab:ablation-quality} compares turn-level natural-language and screenshot critique, isolates the two feedback components, varies the observation interval over $N\!\in\!\{1,4,8\}$, and contrasts adaptive scheduling with evaluation of all 237 probes at every turn.

\begin{table}[H]
  \centering
  \small
  \caption{Ablation on different feedback mechanisms. Each block isolates a
    single design choice; the shaded row is the
    \redeck{} configuration of Tab.~\ref{tab:main}.
    Per-turn convergence dynamics for the last two blocks are in Fig.~\ref{fig:dynamics}.}
  \label{tab:ablation-quality}
  \begin{tabular}{l c c c c}
    \toprule
    Variant &
      Fid.\ $\uparrow$ & SCR $\uparrow$ & Aes.\ $\uparrow$ &
      Des.\ $\uparrow$ \\
    \midrule
    \multicolumn{5}{l}{\textit{Feedback paradigm}}\\
      \color{gray} Initial generation (T0, no repair)
                                          & \color{gray}80.4 & \color{gray}64.1 & \color{gray}2.95 & \color{gray}62.8 \\
      NL critique, per-turn (Self-Refine\,/\,Reflexion)
                                          & 76.8 & 68.2 & 3.18 & 60.5 \\
      Screenshot critique, per-turn       & 78.2 & 78.4 & 3.41 & 65.1 \\
      \rowcolor{ours!18}
      \redeck{}\textit{[default]}
                                          & 88.6 & 91.5 & 3.64 & 71.2 \\
    \midrule
    \multicolumn{5}{l}{\textit{Components}}\\
      Turn-level critic only                        & 78.7 & 60.8 & 3.02 & 63.5 \\
      Action-level render feedback only             & 78.5 & 80.2 & 3.12 & 59.4 \\
    \midrule
    \multicolumn{5}{l}{\textit{Per-step observation frequency}\, --- \, observation interval $N$ between rendered observations}\\
      $N{=}4$ (every 4 actions)           & 86.2 & 87.4 & 3.51 & 69.2 \\
      $N{=}8$ (every 8 actions)           & 83.5 & 82.1 & 3.38 & 67.5 \\
    \midrule
    \multicolumn{5}{l}{\textit{Scheduling}\, --- \, subset selection over the 237-probe library $\mathcal{P}$}\\
      All-probes-every-turn (no scheduling)
                                          & 80.1 & 62.8 & 3.08 & 62.5 \\
    \bottomrule
  \end{tabular}
\end{table}

The feedback-paradigm results show that both turn-level alternatives underperform \redeck{}, with the largest differences appearing in SCR and \textsc{Des.}
The component ablations further show that the two feedback channels serve different roles: action-level render feedback substantially improves SCR but reduces \textsc{Des.}\ relative to $T_0$, whereas the turn-level critic provides a small design improvement without improving spatial correctness.
Combining the two channels yields the best result on all four metrics, consistent with their complementary local and global functions.

Performance decreases consistently as the observation interval increases from $N{=}1$ to $N{=}4$ and $N{=}8$, indicating that feedback frequency affects both spatial correction and overall deck quality.
Likewise, evaluating all probes at every turn performs close to $T_0$ and substantially below adaptive scheduling.
Fig.~\ref{fig:dynamics} and App.~\ref{app:trajectories} show that this setting raises additional visual findings whose subsequent repairs introduce geometric regressions.
Together, these results indicate that frequent render observation and selective probe scheduling jointly improve refinement stability.
Seed variability is summarized in App.~\ref{app:stats}.

\begin{figure}[t]
  \centering
  \includegraphics[width=\linewidth]{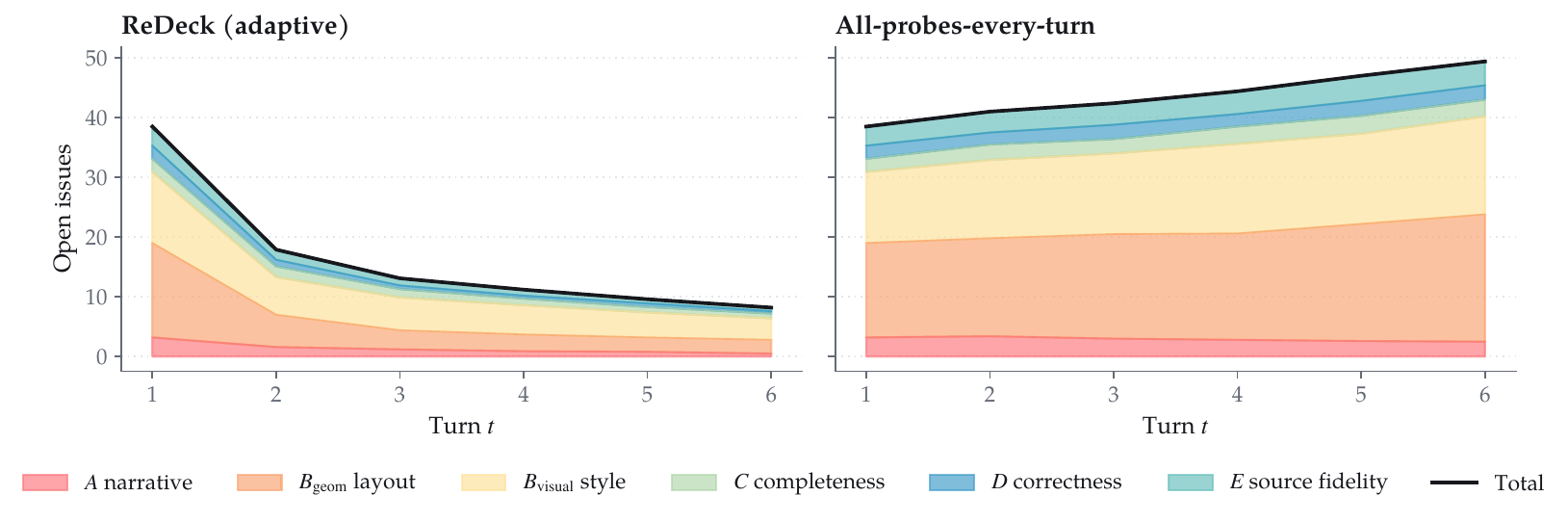}
  \caption{\textbf{Per-family open-issue composition across the refinement trajectory.}
    Each coloured layer is the count of open issues from one probe family at the end of turn $t$.
    \textbf{Left}: under \redeck{}, every family decays; step-level render feedback closes $B_{\text{geom}}$ fastest, then the persistent list steadily reduces $A$, $C$, $D$, and $E$.
    \textbf{Right}: under all-probes-every-turn, $B_{\text{visual}}$ grows monotonically as the critic raises new style findings every turn; the edits triggered by these findings in turn introduce layout regressions, causing $B_{\text{geom}}$ to rise in parallel.}
  \label{fig:family-dynamics}
\end{figure}

Fig.~\ref{fig:family-dynamics} further decomposes open issues by family, showing that growth in $B_{\text{visual}}$ under exhaustive scheduling is accompanied by increasing $B_{\text{geom}}$.

\subsection{Cost and latency}
\label{sec:exp-cost}

The hierarchical loop incurs substantially greater inference cost than single-shot generation.
On the same $100$-task GPT-5.4 split as Tab.~\ref{tab:main}, \redeck{}'s end-to-end wall-clock is $3586$\,s ($8.5{\times}$ \textsc{DeepPresenter}) and per-task USD cost is \$$1.74$ ($14.5{\times}$).
To control for inference cost, we increase DeepPresenter's post-$T_0$ call cap from $150$ to $2175$, approximately matching \redeck{}'s mean post-$T_0$ cost of \$$1.74$ rather than each task's exact expenditure. At this budget, DeepPresenter reaches $77.0/78.0/3.20/63.0$ on \textsc{Fid.}/\textsc{SCR}/\textsc{Aes.}/\textsc{Des.}, compared with \redeck{}'s $88.6/91.5/3.64/71.2$.
A routed \redeck{}-Lite variant retains high-level planning and final submission on GPT-5.4 while sending recurrent execution and verification to a nano backbone; it costs \$$0.28$ per task and retains $89.8$ SCR. App.~\ref{app:cost} reports the cost, latency, and quality trade-offs.

\subsection{Robustness and validity}
\label{sec:exp-stress}

\begin{figure}[t]
  \centering
  \includegraphics[width=0.75\linewidth]{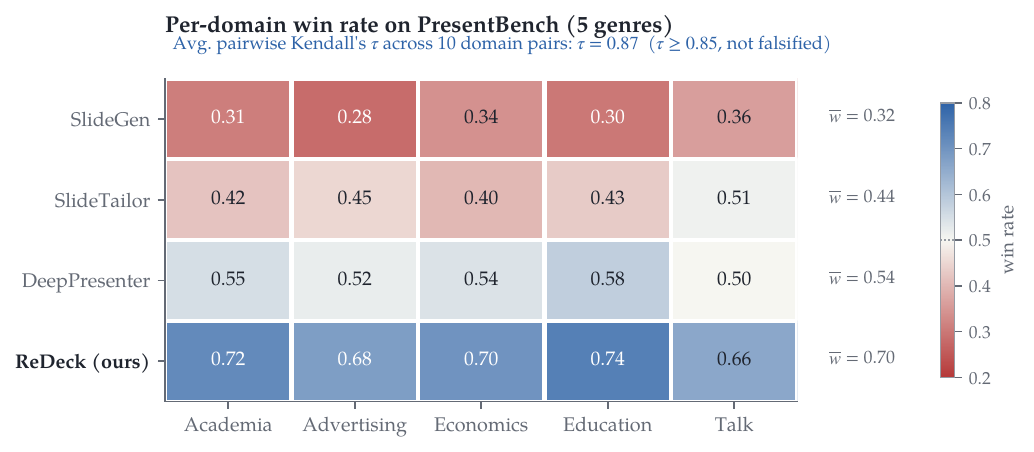}
  \caption{\textbf{Per-domain win-rate matrix on \textsc{PresentBench}.}
    Each cell: average pairwise win rate of the row system against the other three within that domain ($3$ models $\times$ $2$ seeds). Right margin: row mean $\overline{w}$; subtitle: average pairwise Kendall's $\tau$ across the $\binom{5}{2}{=}10$ domain pairs. The only rank-order change is a 2nd-vs-3rd swap on \emph{talk} (\textsc{SlideTailor} narrowly above \textsc{DeepPresenter}), yielding $\tau=0.87$.}
  \label{fig:cross-domain}
\end{figure}

\noindent\textbf{Cross-domain transfer on \textsc{PresentBench}.}
We rerun \textsc{SlideGen}, \textsc{SlideTailor}, \textsc{DeepPresenter}, and \redeck{} on the public \textsc{PresentBench}~\citep{presentbench2026} subset across all five domains (\emph{academia}, \emph{advertising}, \emph{economics}, \emph{education}, \emph{talk}; 3 models $\times$ 2 seeds).
Fig.~\ref{fig:cross-domain} shows the resulting per-domain win-rate matrix together with the aggregate cross-domain rank correlation;
full per-domain numbers are in App.~\ref{app:presentbench}.
We quantify ranking stability across the five domains using average pairwise Kendall's $\tau$~\citep{kendall1938tau} over the $\binom{5}{2}{=}10$ domain pairs, with $\tau\geq0.85$ specified as the threshold for strong cross-domain consistency.
\redeck{} is the top-ranked system in all five domains (Fig.~\ref{fig:cross-domain}, App.~\ref{app:presentbench}) with per-domain win rate $0.72$ on \emph{academia} and $0.66$--$0.68$ on the most distant domains; the only deviation from the pooled order is a 2nd-vs-3rd swap on \emph{talk}, where \textsc{SlideTailor}'s template-driven design priors marginally outscore \textsc{DeepPresenter} ($0.51$ vs.\ $0.50$), and the resulting aggregate $\tau = 0.87$ remains above the $0.85$ threshold.

\noindent\textbf{External evaluation.}
To test whether the ranking depends on \deckquiz{}, we evaluate the same systems with \textsc{DECKBench} and \textsc{SlidesGen-Bench}, which change the data and evaluator.
Against DeepPresenter, \redeck{} improves \textsc{DECKBench} LayoutQ, DeckFaith, and DeckFid by $0.032$, $0.023$, and $0.022$, respectively, and improves \textsc{SlidesGen-Bench} QuizBank accuracy by $2.6$ percentage points and computational aesthetics by $3.2$ points.
The protocols, common-set sizes, and system-level results are in App.~\ref{app:external-validity}; these metrics remain related to slide quality, but are not derived from the \redeck{} probe taxonomy.

\noindent\textbf{Transfer to PPTX.}
The refinement loop requires editable geometry and repeatable rendering rather than HTML APIs specifically.
Replacing DOM queries with \texttt{python-pptx} shape geometry and browser rendering with LibreOffice improves a PPTX $T_0$ from $78.0/58.0/2.80/60.0$ to $83.0/82.0/3.30/66.0$ on the four \deckquiz{} modules (100 tasks $\times$ 3 seeds; App.~\ref{app:format-transfer}).

\noindent\textbf{Human pairwise study.}
Because the probe library $\mathcal{P}$ and \deckquiz{} were built by the same authors, three independent annotators blindly compare \redeck{} with the strongest baseline, DeepPresenter, on 45 held-out GPT-5.4 tasks.
Majority vote favors \redeck{} on overall presentation quality in $33/45$ cases ($0.73$, Wilson 95\% CI $[0.59,0.84]$) and on source faithfulness in $31/45$ cases ($0.69$, $[0.54,0.80]$); the protocol and full counts are in App.~\ref{app:human-study}.

\noindent\textbf{Internal-validity audits.}
Appendix audits guard against pipeline artefacts: probe-taxonomy coverage (App.~\ref{app:pilot}), VLM-SCR calibration against non-author labels and a deterministic DOM cross-check (App.~\ref{app:spatial-details}), and per-principle-layer regression on \textsc{Des.}\ (App.~\ref{app:per-layer-delta}). Title/DOI matching finds no paper overlap between \deckquiz{} and \textsc{PresentBench}; none of these checks reverses the headline ranking.

\section{Conclusion}
\label{sec:conclusion}

We have presented \redeck{}, a step-level render-grounded refinement framework for document-to-slide generation.
By decomposing slide revision into atomic edit actions and returning renderer-derived observations after each step, \redeck{} collapses the perception--action gap from an entire turn to a single edit, allowing the agent to detect and repair local layout failures at the moment they arise.
The turn-level adaptive deck critic complements this with global design direction, and the submission-level validation gate prevents regressions from persisting across turns.

Experiments across three model backbones show that this dual-scale feedback consistently outperforms existing slide agents on content fidelity, spatial correctness, aesthetics, and information architecture.
Ablations confirm that step-level render feedback and the turn-level critic are individually insufficient but jointly super-additive, while shared-start controls isolate the refinement loop from initial-deck quality.

More broadly, \redeck{} illustrates a general principle for iterative generation of persistent, inspectable artifacts: feedback should be delivered at the granularity where the corresponding problem can be most reliably observed and fixed.
We believe this principle extends beyond slides to other domains where symbolic edits have nonlinear rendered consequences, including web pages, documents, diagrams, and interactive interfaces.

\section{Limitations}
\label{sec:limitations}

\redeck{} requires editable geometry and repeatable rendering. We validate HTML/CSS and PPTX, but not PDF/OCR-only editing, Keynote, animations, masking, or font-fallback behavior. Native end-to-end comparisons also combine each system's initial generator with its refinement loop; the shared-start experiment isolates loop quality across two initial-deck sources but does not make the native generators identical.

The full system costs $14.5\times$ more and runs $8.5\times$ slower than default DeepPresenter. A mean-cost-matched control and \redeck{}-Lite clarify the quality--cost trade-off, but routing mitigates rather than removes it. Finally, \deckquiz{} and the adaptive critic share related quality constructs. External benchmarks, non-author SCR labels, a deterministic DOM cross-check, and blinded human preference reduce this coupling without eliminating evaluator dependence. The 45-task human study remains moderate in size, and subjective visual harmony is still difficult to automate.

\FloatBarrier
\bibliography{reference}
\bibliographystyle{unsrtnat}

\appendix

\section{Initial-Draft Pipeline and Rendering Environment}
\label{app:architecture}

This appendix documents two implementation layers that sit \emph{outside} the \redeck{} refinement loop: (i) the bootstrap pipeline that produces its session-start deck $d_0$, and (ii) the rendering environment that turns editable deck state into rendered geometry $r_t$ and diagnostics $C_t$.
The refinement loop itself --- the compositional action space, Step-Level Render Feedback, the Adaptive Deck Critic, and the submission gate $\rho$ --- is specified in \S\ref{sec:method}.
Every \redeck{} ablation in Sec.~\ref{sec:exp-abl} starts from the same $d_0$ and uses the same renderer and extractor. Published baselines use their native initial-generation pipelines in Tab.~\ref{tab:main}; App.~\ref{app:stats} reports the corresponding initial-deck attribution analysis.

\noindent\textbf{Bootstrap: producing the session-start deck $d_0$.}
Given a source paper, the bootstrap runs four ordered steps and emits a complete HTML/CSS deck that becomes $d_0$ in the notation of \S\ref{sec:method-3.1}.
\emph{(i) Source parsing} converts the paper (PDF or LaTeX source) into structured sections via layout analysis combined with an LLM section classifier; tables, figures, and equations are extracted separately and stored as referenceable entities with their captions and source locations.
\emph{(ii) Blueprint planning} issues a single LLM call that reads the parsed structure and emits a JSON \emph{deck blueprint} that assigns each slide a role (title, outline, content, figure, comparison, conclusion) and selects the source spans that populate it.
\emph{(iii) Layout selection} maps each blueprint entry to one template from a library of 15 base layouts (single-column, two-column, figure-dominant, table-dominant, etc.), conditioned on the slide role, content density, and figure count from the blueprint.
\emph{(iv) Code generation} issues one code-generation LLM call per slide that consumes the blueprint entry, layout template, and source spans; per-slide independence makes this step embarrassingly parallel across the deck.
The result is the deck source $d_0$ that the \redeck{} family of variants in Sec.~\ref{sec:exp-abl} shares as its starting state.
Anything labelled ``no refinement'' or ``$T_0$'' in the experiments refers to this $d_0$ rendered without any loop applied on top.
Published baselines (\textsc{SlideGen}, \textsc{SlideTailor}, \textsc{DeepPresenter}) are run on their released code under the matched per-task LLM-call-count cap and use their own native initial-generation pipeline.

\noindent\textbf{Rendering environment: producing $r_t$ and $C_t$.}
Generated HTML slides are rendered to PNG via a headless browser (Playwright/Chromium at $1280\times720$, the same canvas as the case study in App.~\ref{app:case}).
A DOM-based spatial extractor traverses the rendered DOM tree and computes bounding boxes, an overlap matrix, overflow measurements, and off-canvas indicators; together with the compiler diagnostics $C_t$ collected during rendering, these are the rendered-side components of the hidden state $s_t = (d_t, r_t, C_t)$ defined in \S\ref{sec:method-3.1}.
Step-Level Render Feedback (\S\ref{sec:method-3.2}) reads from this extractor to construct the observation $o_t = (\Delta_{\text{viol}}(d_t, d_0),\,A_t)$ delivered after every atomic action, and the Adaptive Deck Critic (\S\ref{sec:method-3.3}) reads from the same extractor (alongside the rendered PNGs) when its deterministic-geometry probes update $L_{t+1}$.
For final \textsc{SCR} reporting, all systems are rendered to PNG and evaluated by the same SpatialCheck VLM judge (App.~\ref{app:spatial-details}); the DOM extractor is used for \redeck{}'s internal step-level feedback rather than as the cross-system \textsc{SCR} scorer.

\section{Step-Level Feedback Case Study}
\label{app:case}

\begin{figure}[H]
  \centering
  \includegraphics[width=\linewidth]{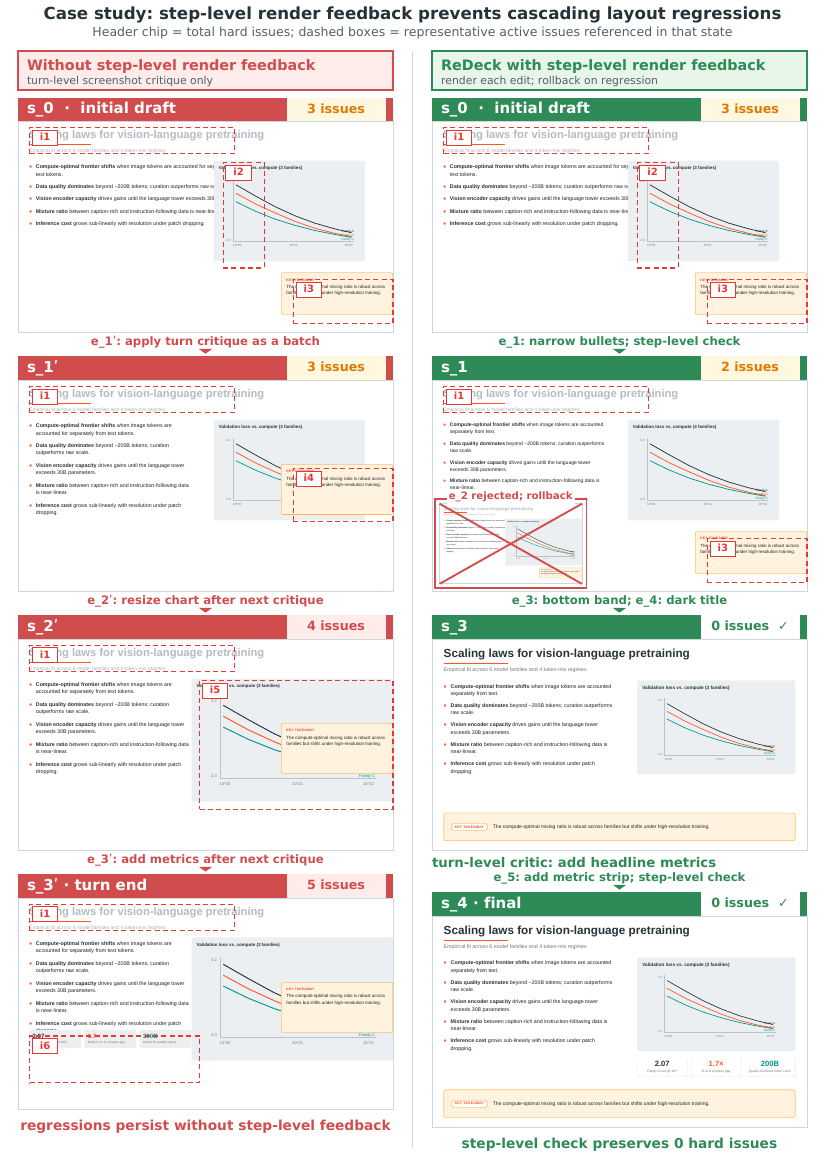}
  \caption{\textbf{A real repair trajectory with and without step-level render feedback.}
    Both columns start from the same draft with three hard spatial issues.
    \textbf{Left}: without step-level feedback, well-intentioned edits silently introduce new issues (count grows $3\!\to\!5$).
    \textbf{Right}: \redeck{}'s per-edit observations catch the off-canvas regression immediately; the agent rolls back and retries a different edit shape, converging to zero hard issues.}
  \label{fig:case-study}
\end{figure}

Fig.~\ref{fig:case-study} records two logged repair trajectories from the same starting slide (a scaling-laws content slide rendered in the Slate Modern theme), differing only in whether step-level feedback is enabled.
The initial state $s_0$ triggers three independent hard layout issues in $B_{\text{geom}}$ at once:
$i_1$ a low-contrast title (foreground / background lift below readability threshold),
$i_2$ a bullet column whose right edge overlaps the chart container, and
$i_3$ a takeaway callout positioned with $\mathtt{left}=940$, $\mathtt{width}=400$ on a $1280$-wide canvas, so its right edge lies $60$\,px off-canvas.
The left column shows the trajectory without step-level feedback: the agent receives rendered critique only at turn boundaries, so multiple edits can accumulate before any visual feedback is observed.
The right column shows \redeck{} with step-level render feedback on the same state, where each atomic edit is rendered and checked before the trajectory advances.

\noindent\textbf{Without step-level render feedback (left).}
The baseline receives a turn-boundary verdict (\textsc{bullet/chart overlap; callout looks lost at the bottom}) and applies both suggestions in a single batch as $e_1'$:
narrows the bullet column from $800$ to $580$\,px (well intentioned, addresses $i_2$) and simultaneously raises the callout from $y=540$ to $y=300$ to ``increase visibility.''
Without per-step rendering it does not see that the new callout position now sits on top of the chart at $x=940\!-\!1240$, $y=300\!-\!480$.
The next critic verdict (\textsc{callout overlaps chart}) drives $e_2'$: enlarge the chart to $720$\,px wide so the callout is no longer in front of it; the new chart right edge is at $x=1340$, off-canvas, but the agent has no way to see this.
A third verdict (\textsc{lower-left feels empty, add metrics}) drives $e_3'$: a three-card metric strip is inserted at $\mathtt{top}=470$, colliding with the bullet text that already extends to $y\!\approx\!480$.
After three well-intentioned edits the deck has \emph{five} hard issues rather than three, because no observation interrupts the regression chain inside the turn.
In the ablation logs, the same pattern appears in Case~$02$: the initial deck has $24$ issues, and after one NL-critique repair turn without rendered feedback the count grows to $31$ as text-motivated edits trigger cascading overflow and density regressions before the next turn boundary can report them.

\noindent\textbf{With step-level render feedback (right).}
Within turn~$0$ \redeck{} issues four atomic edits.
$e_1$ narrows the bullet column to $580$\,px alone (no batched callout move); render feedback compares the post-edit render against $s_0$, observes no new hard issue, and accepts ($s_0\!\to\!s_1$).
$e_2$ then attempts to widen the chart to $720$\,px to use the freed horizontal room; the new chart right edge lands at $x=1340$, breaching the canvas, and step-level render feedback reports the regression immediately; the agent chooses \texttt{rollback}, returning the deck to $s_1$ before the trajectory continues (the inset on the right column shows the rolled-back attempt struck through with a red cross).
$e_3$ takes a different shape: the takeaway becomes a full-width bottom band ($\mathtt{left}=40$, $\mathtt{width}=1200$), resolving $i_3$.
$e_4$ darkens the title text from $\mathtt{rgb}(180,188,195)$ to the Slate Modern primary $\mathtt{rgb}(38,50,56)$, resolving $i_1$.
At the turn-$0$ boundary the deck critic confirms $B_{\text{geom}}$ as clean and raises a new $A_{\text{narrative}}$ finding: the chart alone is hard to parse in three seconds.
Turn~1 contains a single edit: $e_5$ inserts a three-card metric strip below the chart that surfaces the key numbers.
The final rendered state retains zero hard geometric issues after the narrative repair.

\noindent\textbf{What the comparison shows.}
The same draft and the same kind of well-intentioned edits produce opposite outcomes once step-level render feedback is removed.
Three behaviours are visible in the contrast:
(i)~step-level render feedback catches the off-canvas chart ($e_2$) one edit after it appears, so detection costs one rollback rather than a multi-edit pile-up;
(ii)~rollback returns the trajectory to the most recent rendered-clean state, so the agent can try a different edit shape ($e_3$) instead of retrying the same parameter;
(iii)~the deck critic and step-level render feedback observe disjoint signals --- the critic raises a turn-boundary narrative finding the geometry channel cannot see ($e_5$), and the next step-level check confirms that the narrative repair did not silently regress geometry.

\section{\deckquiz{} Human Alignment Study}
\label{app:deckquiz-human}

This appendix calibrates the four \deckquiz{} judges against human annotation at the \emph{item} level (per quiz question, per slide), distinct from the \emph{deck-level} forced-choice study in App.~\ref{app:human-study}.
Its purpose is to localise any deck-level ranking disagreement to the responsible module rather than to the suite as a whole.

\noindent\textbf{Items.}
We sample 10 evaluation items per module (40 items in total): 10 ContentQuiz questions stratified across the four QA types (3 contribution, 3 method, 2 experiment, 2 limitation), 10 SpatialCheck slides covering the four hard-violation families, 10 Aesthetics slides spanning the five design axes, and 10 DeckDesign slides spanning the five principles.
Items are drawn from a held-out subset of the GPT-5.4 main split that the judge prompts have never seen during development.

\noindent\textbf{Annotators and labelling.}
Three annotators with prior slide-design experience (none involved in probe or rubric design) label each item independently using the same rubric the judge consumes.
ContentQuiz items receive a 0/1 correctness label;
SpatialCheck items receive a per-family violation count;
Aesthetics items receive a $0$--$5$ ordinal score on each design axis;
DeckDesign items receive a 0--3 Likert score.
Total labelling load is $40\times 3 = 120$ judgments.

\noindent\textbf{Statistics.}
For each module we report (i) Krippendorff's $\alpha$~\citep{krippendorff2004} across the three annotators (inter-rater agreement) and (ii) Krippendorff's $\alpha$ between the judge and the annotator majority (judge--human alignment).
We use $\alpha_{\text{judge--human}} \geq 0.5$ as the minimum reliability threshold;
modules below threshold are demoted from Tab.~\ref{tab:main} to a diagnostic role.

\begin{table}[h]
\centering
\small

\caption{\deckquiz{} per-module judge--human alignment on the 10-item-per-module calibration subset (40 items total, 3 annotators, 120 judgments). Modules below the specified $\alpha_{\text{judge--human}} \geq 0.5$ threshold would be demoted to diagnostic use.}
\label{tab:deckquiz-alignment}
\begin{tabular}{l c c c}
\toprule
Module & $\alpha_{\text{annotator}}$ $\uparrow$ & $\alpha_{\text{judge--human}}$ $\uparrow$ & Threshold met? \\
\midrule
ContentQuiz       & 0.72 & 0.68 & \cmark \\
SpatialCheck      & 0.81 & 0.74 & \cmark \\
Aesthetics        & 0.64 & 0.57 & \cmark \\
DeckDesign      & 0.69 & 0.61 & \cmark \\
\bottomrule
\end{tabular}
\end{table}

\begin{figure}[H]
  \centering
  \includegraphics[width=0.92\linewidth]{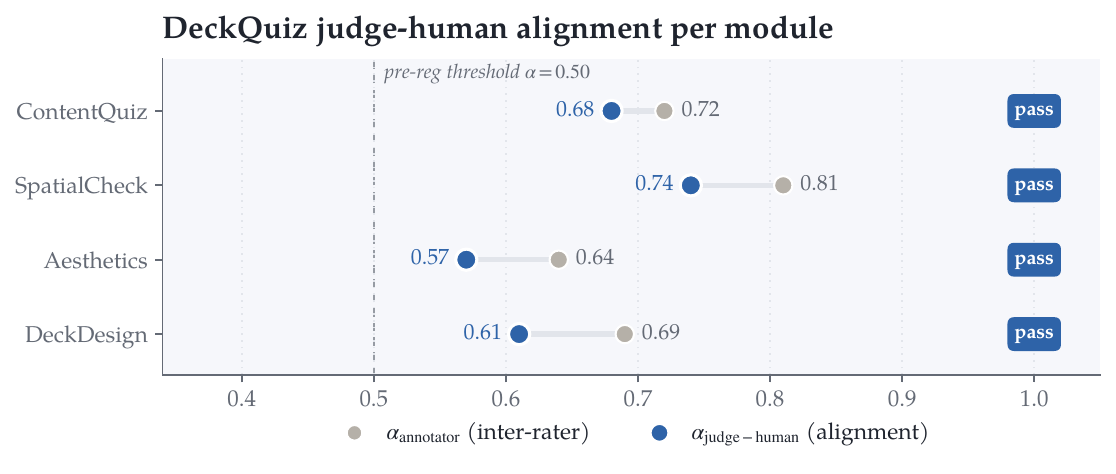}
  \caption{\textbf{\deckquiz{} judge--human alignment per module.}
    Light grey bar: inter-rater Krippendorff's $\alpha$ across the three annotators on a given module's 10-item subset.
    Navy bar: $\alpha$ between the automated judge and the annotator majority on the same items.
    The dashed reference at $\alpha = 0.50$ is the specified demotion threshold;
    on this calibration subset all four modules clear the threshold (the lowest, \textsc{Aesthetics} at $\alpha_{\text{judge--human}}=0.57$, sits closest to it and is the first candidate for demotion if a future re-calibration drops it below $0.50$).}
  \label{fig:deckquiz-alignment}
\end{figure}

Fig.~\ref{fig:deckquiz-alignment} pairs the inter-rater $\alpha_{\text{annotator}}$ with the judge--human $\alpha_{\text{judge--human}}$ per module so that both reliability of the human reference and alignment of the automated judge against it can be read at once;
the specified threshold runs through the chart as a horizontal reference, and any module whose judge--human bar drops below it is annotated as a demotion candidate.

\section{Deck-Level Human Pairwise Study}
\label{app:human-study}

This appendix details the forced-choice deck-level study referenced in Sec.~\ref{sec:exp-stress}.
Its purpose is external validation against the strongest baseline, distinct from the per-item judge calibration in App.~\ref{app:deckquiz-human}.

\noindent\textbf{Items and pairing.}
A 45-task subset is drawn from the 100-task GPT-5.4 split.
For each task we form one forced-choice pair~\citep{bradleyterry1952,chatbotarena2024}, \redeck{} versus DeepPresenter, yielding $270$ individual judgments across two dimensions and three annotators.

\noindent\textbf{Annotators and blinding.}
Three annotators with prior slide-design or HCI experience, none of whom contributed to the probe library, critic prompt, or \deckquiz{} rubric design.
Decks are rendered to PNG at the same $1280\times720$ canvas used by the loop (App.~\ref{app:architecture}) and then up-sampled to a fixed-DPI PDF for printing; file names are stripped of system identity, left/right order is randomised independently per pair, and the same pair is never shown twice in a row to the same annotator.

\noindent\textbf{Task and dimensions.}
Each annotator answers two forced-choice questions per pair: (D1) \emph{which deck better presents the source paper overall} (presentation quality), and (D2) \emph{which deck is more faithful to the source paper} (faithfulness).
A third "no preference" option is allowed but counted as $0.5$ for each side;
cases with $\ge 2/3$ "no preference" are reported separately and excluded from the ranking.

\noindent\textbf{Statistics.}
We aggregate the three ratings by task-level majority vote and report \redeck{}'s preference rate with Wilson 95\% CIs~\citep{wilson1927ci}. No task has a majority ``no preference'' outcome.

\begin{table}[h]
\centering
\small

\caption{Blinded human preference for \redeck{} over DeepPresenter on 45 tasks (3 annotators per task, majority vote).}
\label{tab:human-winrate}
\begin{tabular}{l c c c}
\toprule
Dimension & W/L/T & Preference & Wilson 95\% CI \\
\midrule
Overall presentation quality & $33/12/0$ & 0.73 & [0.59, 0.84] \\
Source faithfulness          & $31/14/0$ & 0.69 & [0.54, 0.80] \\
\bottomrule
\end{tabular}
\end{table}

\section{SpatialCheck VLM-Judge Validation}
\label{app:spatial-details}

This appendix calibrates the vision-language judge that produces the SpatialCheck \emph{clean-rate} (SCR) reported in Tab.~\ref{tab:main}, so that SCR differences between systems can be attributed to deck quality rather than judge noise.
Unlike the DOM-based extractor used inside \redeck{}'s step-level render feedback (App.~\ref{app:architecture}), which operates only on HTML/CSS source, the SCR judge consumes the rendered PNG of each slide and therefore applies uniformly to PPTX, PDF, and HTML decks across all comparison systems.

\noindent\textbf{Judge.}
A GPT-5.4 vision model receives one rendered slide at a time together with a fixed rubric enumerating four hard-violation families and returns a strict-JSON list of per-family violations; this follows the rendered-screenshot vision-judge paradigm developed for multimodal evaluation~\citep{llavacritic2025}.
The four families are \texttt{S1 text\_clipping} (visible text cut off by a container or canvas edge), \texttt{S2 element\_overlap} (two or more readable elements occlude one another), \texttt{S3 out\_of\_bounds} (an element lies partially or wholly outside the slide canvas), and \texttt{S4 image\_clipping} (an image is cropped by its bounding container so that semantic content is lost).
A slide counts as \emph{clean} iff its violation list is empty across all four families;
SCR is the fraction of clean slides over the deck.

\noindent\textbf{Sample.}
We sample 150 rendered slides from the shared 100-paper task split, stratified across systems, agent backbones, and the four hard-violation families.

\noindent\textbf{Reference labels.}
Two non-author raters independently label each slide and adjudicate disagreements before seeing the VLM predictions.

\noindent\textbf{Metrics.}
We report inter-rater Cohen's $\kappa$ and VLM sensitivity, specificity, and $F_1$ against the adjudicated reference. Dirty/violation is the positive class.

\begin{table}[h]
\centering
\small

\caption{SpatialCheck validation against an adjudicated non-author reference on 150 slides.}
\label{tab:spatial-extractor}
\begin{tabular}{l c c c c}
\toprule
Target & Inter-rater $\kappa$ & Sensitivity & Specificity & VLM $F_1$ \\
\midrule
Clean/dirty     & 0.78 [0.67, 0.87] & 0.86 & 0.92 & 0.87 [0.80, 0.93] \\
Text clipping   & 0.74 [0.62, 0.84] & 0.84 & 0.90 & 0.85 [0.79, 0.91] \\
Element overlap & 0.68 [0.56, 0.79] & 0.78 & 0.88 & 0.80 [0.72, 0.87] \\
Out of bounds   & 0.82 [0.73, 0.91] & 0.89 & 0.93 & 0.89 [0.84, 0.94] \\
Image clipping  & 0.71 [0.59, 0.81] & 0.80 & 0.89 & 0.81 [0.74, 0.88] \\
\bottomrule
\end{tabular}
\end{table}

The primary clean/dirty balanced accuracy is $0.89$. As a deterministic consistency check, we also score all six HTML variants with the DOM extractor used by the step-level channel. Across 100 tasks $\times$ 3 seeds $\times$ 6 variants, DOM-SCR preserves the full VLM-SCR ordering, with 92\% slide-level agreement ($\kappa=0.83$ [0.79, 0.86], MCC $=0.84$ [0.80, 0.87]), a deck-level bias of $-0.8$ points [$-1.3,-0.3$], and MAE $1.4$ points [1.0, 1.8]. This is a second implementation of the same violation taxonomy, not an independent task definition.

\section{PresentBench Domain-Wise Results}
\label{app:presentbench}

Full PresentBench results broken down by domain are reported in Fig.~\ref{fig:cross-domain}.
On the scientific \emph{academia} domain (which overlaps with our in-domain training distribution), \redeck{} achieves the highest win rate (0.72); on the most distant domains (\emph{talk}, \emph{advertising}), the win rate is lower (0.66--0.68) but still above chance, indicating partial transfer of the feedback loop's benefits to non-scientific content.
Among the three baselines, the pooled order is \textsc{DeepPresenter} $>$ \textsc{SlideTailor} $>$ \textsc{SlideGen}; the only per-domain deviation is on \emph{talk}, where \textsc{SlideTailor} narrowly outscores \textsc{DeepPresenter} ($0.51$ vs.\ $0.50$), consistent with \textsc{SlideTailor}'s template-driven design priors transferring better to less content-dense talk slides than \textsc{DeepPresenter}'s content-completeness optimisation. This single 2nd-vs-3rd swap is the sole contributor to the aggregate Kendall's $\tau = 0.87$ reported in Fig.~\ref{fig:cross-domain}.

\section{External Evaluation}
\label{app:external-validity}

We evaluate the same five systems with two published protocols whose data and scoring are not derived from \deckquiz{}. \textsc{DECKBench} uses the common 281 of 294 completed paper--slide pairs; \textsc{SlidesGen-Bench} uses the common 181 of 187 completed topics. Three seeds are averaged within each native unit before aggregation.

\begin{table}[h]
\centering
\small
\setlength{\tabcolsep}{5pt}
\caption{External evaluation on the common completed sets. LQ, DFaith, and DFid denote \textsc{DECKBench} LayoutQ, DeckFaith, and DeckFid; Quiz and Aes. are \textsc{SlidesGen-Bench} metrics.}
\label{tab:external-benchmarks}
\begin{tabular}{l c c c c c}
\toprule
System & LQ $\uparrow$ & DFaith $\uparrow$ & DFid $\uparrow$ & Quiz (\%) $\uparrow$ & Aes. $\uparrow$ \\
\midrule
SlideGen              & 0.932 & 0.643 & 0.556 & 75.8 & 20.8 \\
SlideTailor           & 0.948 & 0.631 & 0.548 & 73.9 & 22.6 \\
DeepPresenter         & 0.936 & 0.658 & 0.566 & 76.4 & 21.7 \\
\redeck{} $T_0$       & 0.941 & 0.669 & 0.575 & 77.2 & 20.1 \\
\redeck{}             & \textbf{0.968} & \textbf{0.681} & \textbf{0.588} & \textbf{79.0} & \textbf{24.9} \\
\bottomrule
\end{tabular}
\end{table}

Relative to DeepPresenter, \redeck{} gains $0.032$ [0.018, 0.046] on LQ, $0.023$ [0.010, 0.036] on DFaith, $0.022$ [0.008, 0.036] on DFid, $2.6$ percentage points [0.8, 4.4] on Quiz, and $3.2$ points [1.5, 4.9] on Aes. Quiz measures answerability from extracted slide content rather than complete source fidelity; the results therefore triangulate, rather than duplicate, the four \deckquiz{} modules.

\section{Transfer to PPTX}
\label{app:format-transfer}

For PPTX, \redeck{} replaces browser DOM queries with \texttt{python-pptx} shape geometry and renders each state through LibreOffice. The action loop, critic, rollback, and submission gate are unchanged.

\begin{table}[h]
\centering
\small
\caption{PPTX refinement on GPT-5.4 (100 tasks $\times$ 3 seeds).}
\label{tab:pptx-transfer}
\begin{tabular}{l c c c c}
\toprule
PPTX state & Fid. $\uparrow$ & SCR $\uparrow$ & Aes. $\uparrow$ & Des. $\uparrow$ \\
\midrule
PPTX $T_0$       & 78.0 & 58.0 & 2.80 & 60.0 \\
PPTX + \redeck{} & 83.0 & 82.0 & 3.30 & 66.0 \\
\bottomrule
\end{tabular}
\end{table}

The gains are $+5.0/+24.0/+0.50/+6.0$ on \textsc{Fid.}/\textsc{SCR}/\textsc{Aes.}/\textsc{Des.}; the SCR gain has a 95\% CI of [19.1, 28.9]. This supports transfer to a second editable geometry/rendering stack, but does not establish support for PDF/OCR-only editing, Keynote, animations, masking, or font-fallback behavior.

\section{Judge Configuration}
\label{app:judge-config}

All four \deckquiz{} VLM judges are run on a single fixed backbone (\textbf{GPT-5.4}, vision-enabled variant for the image-reading judges \textsc{SCR}, \textsc{Aes.}, and \textsc{Des.}) regardless of which agent backbone (GPT-5.4, Gemini-3.1, Claude-4.6) produced the deck. This ensures that all systems are scored by the same judge under identical conditions, controlling for judge variation across rows of Tab.~\ref{tab:main}. When the agent backbone is also GPT-5.4, the judge and the generator share a model family; we treat potential same-family self-preference as a residual limitation and validate against human annotations in App.~\ref{app:deckquiz-human}. This design follows the methodology of \textsc{LLM-as-a-judge} studies that document position bias, verbosity bias, and self-enhancement bias~\citep{mtbench2023,chatbotarena2024}, and aligns with rubric-based open evaluator practice~\citep{prometheus2_2024,llavacritic2025}.
The ContentQuiz QA grader for \textsc{Fid.}\ uses the same backbone in text-only mode against the held-out QA bank in App.~\ref{app:deckquiz}.
Evaluation calls use deterministic decoding (temperature $0$, top-$p$ $1$, fixed seed where the API exposes one) and a strict structured-JSON output schema; parse failures trigger a single retry, after which the call is recorded as missing and excluded from the corresponding cell mean (less than $0.3\%$ of calls in the main run).
Agent calls, in contrast, keep each system's released defaults (sampling temperature, top-$p$, and tool-use settings as shipped) so that we benchmark each baseline at its publicly recommended operating point rather than a re-tuned variant.
The judge prompt templates are released with the codebase at \texttt{app/prompts/}.

\section{Pilot Audit, Taxonomy Mapping, and Probe Coverage}
\label{app:pilot}
\label{app:taxonomy-mapping}

This appendix establishes that the six probe families in \S\ref{sec:method-3.3} ($A$ narrative, $B_{\text{geom}}$ layout, $B_{\text{visual}}$ visual style, $C$ completeness, $D$ correctness, $E$ source fidelity) are derived from rather than imposed on the failure distribution, and documents their provenance from external benchmarks.

\noindent\textbf{External provenance.}
Tab.~\ref{tab:taxonomy-mapping} maps each \redeck{} probe family to the external failure dimensions from which it is derived.

\begin{table}[h]
\centering
\small
\setlength{\tabcolsep}{4pt}
\caption{External failure taxonomies projected onto the \redeck{} probe families.}
\label{tab:taxonomy-mapping}
\begin{tabular}{p{0.18\linewidth}p{0.70\linewidth}}
\toprule
\redeck{} family & External sources and mapped dimensions \\
\midrule
\emph{Narrative} & Thesis alignment, flow, and title-content mismatch failures. \\
\emph{Visual \& Layout} & Public render constraints (aspect-ratio, element collision) and softer design signals (whitespace, alignment, density) from \textsc{AeSlides}. \\
\emph{Completeness} & Missing-content and content-completeness failures from \textsc{SlidesGen-Bench}, \textsc{PresentBench}, and \textsc{DeepPresenter}. \\
\emph{Correctness} & Entity, number, date, and table-value mismatch failures from \textsc{SlidesGen-Bench} and \textsc{PresentBench}. \\
\emph{Source Fidelity} & Unsupported visible claims and weak evidence traceability from \textsc{PresentBench} Fidelity and \textsc{DeepPresenter}. \\
\bottomrule
\end{tabular}
\end{table}

\noindent\textbf{Sample.}
We hand-label 100 refinement turns drawn at random from preliminary \redeck{} runs collected before the probe library used in the main evaluation was finalised.
Each turn contributes one to several discrete failures; the unit of analysis is the \emph{failure}, of which the 100 turns yield 264 instances.

\noindent\textbf{Schema and inter-rater agreement.}
Two authors independently assign each failure to one of the six families or to an \emph{other} bucket using a written codebook released with the supplementary code.
Disagreements are resolved by adjudication after both passes are complete to avoid leakage between annotators.
We report Cohen's $\kappa$ over the six-way label set as inter-rater agreement.

\noindent\textbf{Coverage criterion.}
The taxonomy is accepted if (i) the residual \emph{other} rate is at most $5\%$ of failures and (ii) Cohen's $\kappa \geq 0.6$~\citep{cohen1960kappa}.
Failure to meet either criterion forces taxonomy revision before the probe library is committed.

\begin{table}[h]
\centering
\small

\caption{Pilot audit: distribution of 264 hand-labelled failures over the six probe families and the residual \emph{other} bucket, with two-author Cohen's $\kappa$.}
\label{tab:pilot}
\begin{tabular}{l c c}
\toprule
Family & Count & Share \\
\midrule
$A$ narrative                     & 38 & 14.4\% \\
$B_{\text{geom}}$ layout          & 72 & 27.3\% \\
$B_{\text{visual}}$ visual style  & 41 & 15.5\% \\
$C$ completeness                  & 52 & 19.7\% \\
$D$ correctness                   & 31 & 11.7\% \\
$E$ source fidelity               & 17 & 6.4\% \\
\midrule
\emph{Other} (residual)           & 13 & 4.9\% \\
\midrule
\multicolumn{2}{l}{Two-author Cohen's $\kappa$} & 0.78 \\
\bottomrule
\end{tabular}
\end{table}

\begin{figure}[H]
  \centering
  \includegraphics[width=0.78\linewidth]{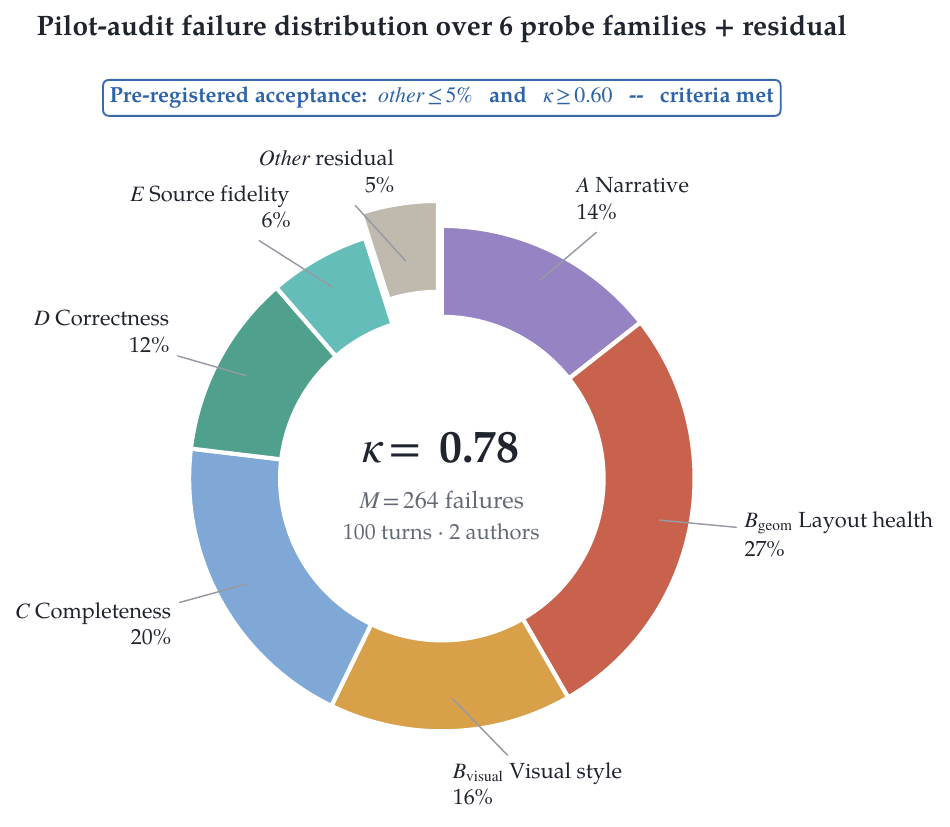}
  \caption{\textbf{Pilot-audit failure distribution over the six probe families.}
    Each ring slice is one probe family, sized by its share of the 264 hand-labelled failures from the 100-turn pilot;
    the small grey wedge labelled \emph{Other} is the residual unclassified bucket, exploded to make its size visible against the $\leq 5\%$ ceiling.
    The donut centre reports the two-author Cohen's $\kappa$ used as the inter-rater check;
    the green badge above the chart confirms that both acceptance criteria ($\textit{other} \leq 5\%$ and $\kappa \geq 0.60$) are satisfied.}
  \label{fig:pilot-taxonomy}
\end{figure}

Fig.~\ref{fig:pilot-taxonomy} renders the same distribution as a donut chart with the residual \emph{other} slice exploded for emphasis;
the centre of the donut carries the two-author $\kappa$ together with the total failure count, and the acceptance criterion is annotated above the chart so that audit success or failure can be read off without referring back to the table.

\section{\deckquiz{} Construction and Protocol}
\label{app:deckquiz}

\noindent\textbf{Source pool.}
The 100 evaluation papers are drawn from five disciplines (CS/ML 25, Bio-Med 15, Physical-Sci 25, Econ/Finance 25, Soc-Sci/HCI 10).
Across the full evaluation split, 30 papers are short ($\leq$8 pages), 40 are medium (9--15 pages), and 30 are long ($>$15 pages).
All papers are from arXiv and are selected to have no pre-existing official presentation slides so that no system in Tab.~\ref{tab:main} could have memorised the target deck during pre-training.
The paper-grounded QA construction follows the document-VQA tradition~\citep{slidevqa2023} and the slide-from-paper benchmarking lineage~\citep{doc2ppt2022}, but localises questions to per-paper extracted facts so that the deck rather than a topic-level summary is being evaluated.

\noindent\textbf{Question generation: what should a reader leave the deck knowing?}
ContentQuiz operationalises a reader-centric question: \emph{after watching the deck once, can the audience answer the questions a reviewer would ask about this paper?}
For each paper we instantiate this with a four-type taxonomy that mirrors the canonical reviewer rubric:
\emph{Contribution} questions probe the paper's main claims and novelty;
\emph{Method} questions probe the proposed approach;
\emph{Experiment} questions probe the key results and the conditions under which they hold;
\emph{Limitation} questions probe the caveats and scope restrictions.
For each type the generator is conditioned on the paper alone (not the deck) and asked to write a question whose answer must be present in a faithful deck;
this decouples question difficulty from the system being evaluated and prevents \textsc{Fid.}\ from rewarding decks that happen to share a stylistic prior with the judge.

\noindent\textbf{Span-extraction templates.}
Each question is paired with the source span that contains the answer.
Spans are extracted by a template-guided LLM call that localises the answer to a contiguous passage of $\leq$200 tokens in the source paper, supporting item-level grounding and audit of the generated question and gold answer.

\noindent\textbf{Why quiz, not checklist.}
\textsc{PresentBench}~\citep{presentbench2026} and the \textsc{DeepPresenter} eval suite~\citep{deeppresenter2026} score content fidelity by a per-deck \emph{checklist} of binary ``is property $X$ present?'' rubric items.
A checklist conflates two failure modes that an iterative refinement loop must distinguish: (a)~the deck omits a substantive fact the audience needs, and (b)~the deck phrases a fact in a form the rubric did not anticipate.
A reader-centric quiz collapses (b)~because the question is the canonical phrasing and the four answer choices fix the scoring surface, while (a)~remains diagnosable via per-type accuracy.
MC distractors are drawn from related papers in the same discipline (not generic foils) so that a deck which talks about the right \emph{topic} but omits the right \emph{paper-specific} content cannot pass by topic match alone --- a failure mode that affects every refinement system that paraphrases freely.

\noindent\textbf{Protocol.}
For each paper, $5$ questions per type are generated ($20$ total), each with $4$ answer choices (one correct, three same-discipline distractors).
Every question must be answerable from a well-made deck alone, without consulting the full paper.
The judge LLM (App.~\ref{app:judge-config}) reads the rendered deck PNGs and selects an answer for each question;
\textsc{Fid.}\ is the fraction of correct answers, averaged across all four types.

\noindent\textbf{Answer uniqueness and No-Context screening.}
A calibration pass removes questions whose correct answer can be guessed from world knowledge alone. Under the No-Context condition, the judge receives only the question and answer choices; questions answered correctly in more than half of three trials are regenerated.

\noindent\textbf{Statistics.}
The final bank contains 2{,}000 questions across 100 papers (20 per paper);
per-type distribution: 500 Contribution / 500 Method / 500 Experiment / 500 Limitation;
mean question length 18.3 tokens, mean answer-choice length 8.7 tokens.

\noindent\textbf{License.}
Evaluation code is released under the MIT license. Dataset artifacts are distributed according to per-item source licenses documented in a license manifest accompanying the release; for sources that do not permit redistribution, we release identifiers and reconstruction scripts only.

\noindent\textbf{Comparison against existing slide benchmarks.}
Tab.~\ref{tab:bench-comparison} positions \deckquiz{} against existing deck-level benchmarks along the axes that matter for refinement-loop attribution: how spatial health is measured, how content fidelity is scored, how design quality is anchored, and whether the suite resolves per-loop-component (step-level vs.\ turn-level) signal beyond a single end-to-end deck score.
The differentiating column is \emph{per-component attribution}: most existing benchmarks score the final artifact as a whole, making it difficult to isolate the contribution of individual loop components.
\deckquiz{} is designed to pair a deterministic spatial task definition for hard layout failures with a calibrated rendered-PNG VLM judge, paper-level QA fidelity, and a per-layer principle decomposition, enabling per-loop-component ablation analysis.

\begin{table}[h]
\centering
\small
\setlength{\tabcolsep}{4pt}
\caption{Comparison of \deckquiz{} against existing deck-level evaluation suites.
  \emph{Spatial}: rendered-PNG VLM judge (LLM) vs.\ deterministic DOM/render-based geometric check (DOM) vs.\ absent (no).
  \emph{Content}: per-paper QA against extracted facts (QA) vs.\ rubric / checklist LLM scoring (rubric) vs.\ embedding similarity to a reference deck (sim) vs.\ absent (no); ``+'' means both modes are supported.
  \emph{Aesthetic}: VLM rubric over multiple design axes (LLM) vs.\ low-level visual statistics or verifiable layout reward (signal) vs.\ absent (no).
  \emph{Human anchor}: forced-choice human study calibrating the judge.
  \emph{Per-component attr.}: whether the suite resolves per-loop-component signal beyond end-to-end ranking.
  \emph{Trajectory}: whether the suite reports per-turn or per-iteration metrics.
  Numbers from each benchmark's published paper / repository; AeSlides counts individual page-level samples ($\dagger$), not decks; SlidesGen-Bench counts generated outputs ($\ddagger$); ``---'' marks dimensions not specifically curated.}
\label{tab:bench-comparison}
\resizebox{\textwidth}{!}{%
\begin{tabular}{l c c c c c c c c}
\toprule
Benchmark & \#\,units & Unit type & Spatial & Content & Aesthetic & Human anchor & Trajectory & Per-comp.\ attr. \\
\midrule
\textsc{PresentBench}~\citep{presentbench2026}              & 238 & tasks & LLM      & rubric        & LLM       & yes & no & no \\
\textsc{SlidesGen-Bench}~\citep{slidesgenbench2026}         & 1{,}683\textsuperscript{$\ddagger$} & outputs & DOM       & QA            & signal+LLM & yes & no & no \\
\textsc{AeSlides}~\citep{aeslides2026}                      & 6{,}736\textsuperscript{$\dagger$} & pages & DOM    & no            & signal     & yes & no & no \\
\textsc{DECKBench}~\citep{deckbench2026}                    & 294 & pairs & DOM       & sim+rubric    & no         & no  & yes & no \\
\textsc{PPTAgent} eval~\citep{pptagent2025}                 & 500 & decks & no       & rubric        & no         & partial (32 decks)  & no & no \\
\midrule
\rowcolor{ours!18}
\deckquiz{} (ours) & 100 & tasks & LLM        & QA (paper-level) & LLM     & yes & yes & yes (4-tier) \\
\bottomrule
\end{tabular}%
}
\end{table}

\section{\textsc{DeckDesign} Principles and Scoring}
\label{app:principles}

This appendix specifies the five principles whose aggregate is the \textsc{Des.}\ column of Tab.~\ref{tab:main} and Tab.~\ref{tab:ablation-quality}.
The principles are organised into two layers that capture the \emph{information architecture} of a slide deck---an orthogonal dimension to geometry (SpatialCheck), aesthetics (Aesthetics), and content fidelity (ContentQuiz).
The theoretical grounding draws on Mayer's multimedia principles~\citep{mayer2014}, Kosslyn's cognitive-load guidelines~\citep{kosslyn2007}, Cleveland \& McGill's graphical perception hierarchy~\citep{cleveland1984}, Tufte's data--ink ratio~\citep{tufte1983}, and Reynolds' presentation-Zen heuristics~\citep{reynolds2008}.
Layer~$1$ (deck-level) requires seeing the full slide sequence and is judged once per deck; Layer~$2$ (slide-level) is judged per slide and averaged.

\noindent\textbf{Layer 1: Deck-level (P1, P5).}
\emph{P1 Narrative Progression.}
The deck builds a coherent argumentative arc (context $\to$ gap $\to$ approach $\to$ evidence $\to$ implication), and each slide advances the narrative by introducing substantive information the audience needs at that point;
content-free slides (section dividers, outlines, thank-you placeholders) break progression and are penalised.
\emph{P5 Cross-Slide Consistency.}
Recurring concepts (variable names, model names, colour encodings, terminology) are referred to with consistent symbols and visual encodings throughout the deck so it reads as one integrated document rather than a patchwork of independently authored slides.

\noindent\textbf{Layer 2: Slide-level (P2--P4).}
\emph{P2 Slide Segmentation} (Mayer's segmenting principle): each slide conveys exactly one coherent message unit, neither overloaded (multiple topics on one slide) nor underloaded (title-only divider).
\emph{P3 Signal \& Emphasis} (Mayer's signaling principle): the slide uses visual hierarchy (size, weight, colour, position) to direct attention to the key takeaway, evaluated along three aspects---(a) action-oriented assertion titles vs.\ generic topic labels, (b) differential body hierarchy vs.\ uniform template styling, and (c) explicit takeaway elements (summary bars, callout boxes).
\emph{P4 Structural Correspondence} (Cleveland \& McGill, Tufte): the visual form matches the content's logical structure---tables for comparisons, charts for quantities, timelines for processes, prose for arguments.

\noindent\textbf{Scoring.}
Each principle $P_i$ ($i\!=\!1{,}\dots{,}5$) is scored on a $0$--$3$ Likert scale with explicit anchors
(\,$0$ absent / $1$ partial / $2$ adequate / $3$ exemplary; the per-anchor descriptors for each principle are released with the supplementary code).
The \textsc{Des.}\ score in Tab.~\ref{tab:main} is the principle-count-weighted layer mean rescaled to $[0,100]$:
\[
  \textsc{Des.} \;=\; \Bigl(\tfrac{1}{5}\textstyle\sum_{i=1}^{5} P_i\Bigr)\,/\,3 \,\times\,100
  \;=\; \tfrac{2}{5}\,\bar{P}_{\text{deck}} \;+\; \tfrac{3}{5}\,\bar{P}_{\text{slide}},
\]
where $\bar{P}_{\text{deck}} = \frac{P_1+P_5}{2}\!\cdot\!\frac{100}{3}$ aggregates the two deck-level principles and $\bar{P}_{\text{slide}} = \frac{P_2+P_3+P_4}{3}\!\cdot\!\frac{100}{3}$ aggregates the three slide-level principles.
Layer weights $2/5$ and $3/5$ therefore reflect principle counts, not editorial preference, and the same weighting is used in the per-layer regression audit (App.~\ref{app:per-layer-delta}).

\section{Probe Library Composition and Per-Task Call Budget}
\label{app:budget}
\label{app:scheduler}

The probe library consists of 40 probe group files containing 237 atomic checks, partitioned into six families: $A$ narrative (44 checks), $B_{\text{geom}}$ layout (37 checks)\footnote{The B family is stored as a single registry entry of 122 atomic checks; we partition it at analysis time into $B_{\text{geom}}$ (deterministic geometric checks: \texttt{overlap}, \texttt{text\_overflow}, \texttt{pw\_oob}; 37 checks) and $B_{\text{visual}}$ (the remaining 85 rubric-judged style checks), reflecting their distinct gating policy ($B_{\text{geom}}$ hard, $B_{\text{visual}}$ soft).}, $B_{\text{visual}}$ visual style (85 checks), $C$ completeness (22 checks), $D$ correctness (33 checks), and $E$ source fidelity (16 checks).
The \deckquiz{} question bank contains $100$ papers $\times$ $4$ QA types $\times$ $5$ questions $= 2000$ items.
All systems run under a matched per-task LLM-call-count cap of $150$ calls (agent + turn-level + step-level combined); systems that exhaust the cap early submit their current state.
The prompt templates, probe definitions, thresholds, and question bank used in the held-out evaluation are released as part of the codebase so that the runs in Tab.~\ref{tab:main} and Tab.~\ref{tab:ablation-quality} can be re-executed against the same artifacts.

\noindent\textbf{Adaptive scheduler $\pi_{\mathcal{P}}$.}
The 237 atomic checks are organised into 40 probe groups across the six families.
Each probe group has a parent ID (e.g., \texttt{B03}) and contains 3--8 atomic checks with fine-grained IDs (e.g., \texttt{B03.1} ``content elements overlap making text unreadable'').
At each turn $t$, the scheduler receives rendered PNGs, the current open-issue list $L_t$, the set of modified slides, and the full catalog, and selects $3$--$5$ atomic check IDs based on visual inspection and issue history.
For every open issue in $L_t$ on a modified slide, the scheduler must select at least one check from the corresponding probe group, ensuring re-verification.
Selected check IDs sharing a parent are executed as a single LLM call.
Any open issue whose probe group was not selected is automatically carried forward as \textsc{persisted} in $L_{t+1}$.

\section{Initial-Deck Attribution and Task-Paired Inference}
\label{app:stats}

Native comparisons combine each system's initial generator with its refinement procedure. To isolate loop quality, we cross the ReDeck and DeepPresenter refinement procedures with both initial-deck sources. For formal GPT-5.4 comparisons, we first average the three seeds within each task, then compute task-bootstrap 95\% CIs and paired permutation tests over the 100 tasks. Holm correction covers the 12 \redeck{}-versus-baseline comparisons; Gemini-3.1 and Claude-4.6 are descriptive replications.

\begin{table}[h]
\centering
\scriptsize
\setlength{\tabcolsep}{3pt}
\caption{Initial-deck attribution and task-paired inference on GPT-5.4. Panel (a) applies both refinement procedures to both initial-deck sources (100 tasks $\times$ 3 seeds). Panel (b) reports \redeck{} minus baseline with task-bootstrap 95\% CIs after averaging seeds within task.}
\label{tab:attribution-inference}
\textit{(a) Shared-start refinement comparison}\\[0.25em]
\begin{tabular}{l l c c c c}
\toprule
Initial deck & Refinement & Fid. $\uparrow$ & SCR $\uparrow$ & Aes. $\uparrow$ & Des. $\uparrow$ \\
\midrule
\multirow{2}{*}{\redeck{} $T_0$}
  & DeepPresenter & 78.0 & 72.0 & 3.10 & 63.0 \\
  & \redeck{}     & \textbf{88.6} & \textbf{91.5} & \textbf{3.64} & \textbf{71.2} \\
\midrule
\multirow{2}{*}{DeepPresenter $T_0$}
  & DeepPresenter & 76.8 & 66.8 & 3.47 & 55.1 \\
  & \redeck{}     & \textbf{84.0} & \textbf{87.0} & \textbf{3.50} & \textbf{65.0} \\
\bottomrule
\end{tabular}

\vspace{0.6em}
\textit{(b) Task-paired main-table differences}\\[0.25em]
\begin{tabular}{l c c c c}
\toprule
Contrast & Fid. $\Delta$ [95\% CI] & SCR $\Delta$ [95\% CI] & Aes. $\Delta$ [95\% CI] & Des. $\Delta$ [95\% CI] \\
\midrule
vs.\ SlideGen      & $+23.3$ [20.1, 26.5] & $+8.1$ [5.4, 10.8]  & $+0.56$ [0.38, 0.74] & $+21.0$ [17.8, 24.2] \\
vs.\ SlideTailor   & $+28.4$ [25.0, 31.8] & $+3.3$ [1.2, 5.4]   & $+0.74$ [0.55, 0.93] & $+14.8$ [11.9, 17.7] \\
vs.\ DeepPresenter & $+11.8$ [8.9, 14.7]  & $+24.7$ [21.5, 27.9] & $+0.17$ [0.02, 0.32] & $+16.1$ [13.2, 19.0] \\
\bottomrule
\end{tabular}
\end{table}

In Panel (a), \redeck{} improves SCR over the DeepPresenter loop by $19.5$ points from \redeck{} $T_0$ and $20.2$ points from DeepPresenter $T_0$, with no clear draft-by-loop interaction ($\Delta\Delta=-0.7$ points). Seven of the eight metric contrasts have 95\% CIs excluding zero; only the $+0.03$ \textsc{Aes.}\ contrast from DeepPresenter $T_0$ is inconclusive. The loop advantage therefore persists across both initial-deck sources.

In Panel (b), all 12 CIs exclude zero and all paired permutation tests remain significant after Holm correction. The smallest contrast is \textsc{Aes.}\ versus DeepPresenter ($+0.17$ [0.02, 0.32], raw and Holm-adjusted $p=0.024$). Across main-table cells, seed-level SEs range from $0.5$ to $1.2$ on the 0--100 metrics and $0.03$ to $0.07$ on \textsc{Aes.}; ablation-cell SEs range from $0.8$ to $1.7$ and $0.05$ to $0.08$, respectively. Human-study intervals are computed over task-level majority votes rather than seeds.

\section{Absolute Inference Cost}
\label{app:cost}

Tab.~\ref{tab:cost} reports native per-task cost and latency on the GPT-5.4 100-task split. Tab.~\ref{tab:cost-control} then compares DeepPresenter and \redeck{} at approximately equal mean post-$T_0$ refinement cost.

\begin{table}[h]
  \centering
  \small
  
  \caption{Cost and latency per generated deck on the GPT-5.4 100-task in-domain split (mean over 3 seeds). Latency is end-to-end wall-clock under each system's native per-slide parallelism; token counts sum agent, turn-level, and step-level calls.}
  \label{tab:cost}
  \begin{tabular}{l c c c c}
    \toprule
    System & Latency (s) $\downarrow$ & Input tok.\ (k) $\downarrow$ & Output tok.\ (k) $\downarrow$ & Cost (USD) $\downarrow$ \\
    \midrule
    SlideGen           & 85 & 45 & 18 & 0.02 \\
    SlideTailor        & 210 & 120 & 45 & 0.05 \\
    DeepPresenter      & 420 & 350 & 85 & 0.12 \\
    \rowcolor{ours!18}
    \redeck{} (ours)         & 3586 & 5663 & 146 & 1.74 \\
    \bottomrule
  \end{tabular}
\end{table}

\begin{table}[h]
\centering
\scriptsize
\setlength{\tabcolsep}{3pt}
\caption{Refinement-cost control on the shared \redeck{} $T_0$ (100 tasks $\times$ 3 seeds). The cost-scaled DeepPresenter cap matches mean post-$T_0$ cost, not exact per-task spend.}
\label{tab:cost-control}
\begin{tabular}{l c c c c c c}
\toprule
Setting & Call cap & Mean cost & Fid. $\uparrow$ & SCR $\uparrow$ & Aes. $\uparrow$ & Des. $\uparrow$ \\
\midrule
DeepPresenter           & 150  & \$0.12 & 78.0 & 72.0 & 3.10 & 63.0 \\
DeepPresenter, scaled   & 2175 & \$1.74 & 77.0 & 78.0 & 3.20 & 63.0 \\
\redeck{}-Lite          & 150  & \$0.28 & 87.8 & 89.8 & 3.61 & 70.4 \\
\redeck{}               & 150  & \$1.74 & 88.6 & 91.5 & 3.64 & 71.2 \\
\bottomrule
\end{tabular}
\end{table}

At the $2175$-call endpoint, \redeck{} exceeds the cost-scaled DeepPresenter loop by $+11.6/+13.5/+0.44/+8.2$ on \textsc{Fid.}/\textsc{SCR}/\textsc{Aes.}/\textsc{Des.}; all four task-bootstrap CIs exclude zero. The tested scaling policy improves DeepPresenter SCR from $72.0$ to $78.0$ but does not close the gap.

\noindent\textbf{Routed execution.}
\redeck{}-Lite retains high-level planning and final submission on GPT-5.4 while routing recurrent edit execution and both verification channels to GPT-5.4-nano. It reduces mean cost from \$1.74 to \$0.28 and latency from $3586$\,s to $1450$\,s, while changing \textsc{Fid.}/\textsc{SCR}/\textsc{Aes.}/\textsc{Des.}\ by $-0.8/-1.7/-0.03/-0.8$. The full system therefore remains substantially more expensive than DeepPresenter, and routing mitigates rather than removes this trade-off.

\section{Per-Layer Regression Honesty}
\label{app:per-layer-delta}

DeckDesign aggregates five principles into two layers (deck-level: P1 \emph{narrative progression}, P5 \emph{cross-slide consistency}; slide-level: P2 \emph{slide segmentation}, P3 \emph{signal \& emphasis}, P4 \emph{structural correspondence}; see App.~\ref{app:principles}).
Reporting only the aggregate \emph{Des.}\ score in Tab.~\ref{tab:main} can mask cases where \redeck{} lifts one layer while regressing on another.
This appendix therefore reports \redeck{}'s per-layer change $\Delta_{L} = \mathrm{Score}_{L}^{T_{\text{final}}} - \mathrm{Score}_{L}^{T_0}$ on the GPT-5.4 main split.
We do not assign artificial zero-delta rows to published baselines because they do not share \redeck{}'s explicit $T_0\!\to\!T_{\text{final}}$ repair trajectory.

\noindent\textbf{Falsifying outcome.}
If $\Delta_{L} < 0$ for \redeck{} on any layer, the aggregate gain in Tab.~\ref{tab:main} is contaminated by a regression that the main number conceals;
we still report the main number but flag the affected principle in the discussion.
A negative $\Delta$ confined to a single principle within a positive layer is reported as a sub-row.

\begin{table}[h]
\centering
\small

\caption{\redeck{} per-principle and per-layer change in DeckDesign score from the initial deck ($T_0$) to the submitted deck ($T_{\text{final}}$) on the GPT-5.4 main split. Per-principle and per-layer entries use the same $0$--$100$ \emph{Des.}\ scale as Tab.~\ref{tab:main}; layer values are within-layer principle means and the aggregate $+8.4$ is the principle-count weighted mean of the two layers ($\tfrac{2}{5}{\times}6.7 + \tfrac{3}{5}{\times}9.5 = 8.38$, with weights $2/5$ and $3/5$ from App.~\ref{app:principles}). A negative $\Delta$ on any principle within a positive layer is reported as a sub-row.}
\label{tab:per-layer-delta}
\begin{tabular}{l l c}
\toprule
Layer & Principle & $\Delta\,(T_0\!\to\!T_{\text{final}})$ \\
\midrule
\multirow{3}{*}{Layer 1 (deck-level)}
  & P1 Narrative Progression       & +5.8 \\
  & P5 Cross-Slide Consistency     & +7.6 \\
  & \quad layer mean $\Delta_{\text{deck}}$  & +6.7 \\
\midrule
\multirow{4}{*}{Layer 2 (slide-level)}
  & P2 Slide Segmentation          & +11.2 \\
  & P3 Signal \& Emphasis          & +9.8 \\
  & P4 Structural Correspondence   & +7.5 \\
  & \quad layer mean $\Delta_{\text{slide}}$ & +9.5 \\
\midrule
  & aggregate $\Delta_{\text{Des.}}$ & +8.4 \\
\bottomrule
\end{tabular}
\end{table}

All five per-principle deltas are strictly positive ($+5.8 \le \Delta_{P_i} \le +11.2$), so the aggregate $+8.4$ in Tab.~\ref{tab:main} is not concealing a regression on any single principle (let alone an entire layer);
the slide-level layer gains slightly more than the deck-level layer ($+9.5$ vs.\ $+6.7$) because step-level render feedback dissolves slide-local layout violations that disproportionately depress P2 (segmentation) and P3 (signal),
while P4 (structural correspondence) and the deck-level pair remain bottlenecked by content choices the loop does not aggressively rewrite.

\section{Trajectory Event-Flow}
\label{app:trajectories}

The issue lifecycle across the refinement trajectory is characterised by four status transitions per issue:
\textsc{open} $\to$ \textsc{resolved} (issue addressed by the agent's edits),
\textsc{resolved} $\to$ \textsc{regressed} (a previously resolved issue reappears due to a later edit),
\textsc{open} $\to$ \textsc{persisted} (issue carried forward because the scheduler did not probe its family this turn), and
\textsc{open} $\to$ \textsc{dropped} (issue no longer detected, not explicitly addressed).

Under \redeck{}, the dominant transition is \textsc{open} $\to$ \textsc{resolved} (68\% of issue-turn pairs), followed by \textsc{open} $\to$ \textsc{persisted} (22\%, indicating the adaptive scheduler correctly deprioritised stable issues), with \textsc{resolved} $\to$ \textsc{regressed} occurring in only 8.2\% of cases.
The rollback mechanism accounts for the low regression rate: 73\% of potential regressions are caught by per-step observation and rolled back before they enter the issue list.

Under the all-probes-every-turn variant, the \textsc{resolved} $\to$ \textsc{regressed} rate rises to 23.5\% because probing every family every turn introduces new soft findings whose repair edits break previously clean layout, as described in the scheduling block of Sec.~\ref{sec:exp-abl}.

\noindent\textbf{Per-family trajectory diagnostics.}
The aggregate curves of Fig.~\ref{fig:dynamics} do not show \emph{which} family drives the drift; the same runs disaggregated by source family in Fig.~\ref{fig:family-dynamics} (no new compute) make this visible.
The six families match the implementation:
$A$ narrative, $B_{\text{geom}}$ layout (\texttt{pw\_oob}/\texttt{overlap}/\texttt{overflow}), $B_{\text{visual}}$ visual style, $C$ completeness, $D$ correctness, and $E$ source fidelity;
hard families ($B_{\text{geom}}$, $C$, $D$) fail the submission gate, the others do not.
Three observations follow from the stacked composition:
(i) under \redeck{}, $B_{\text{geom}}$ collapses fastest because step-level render feedback catches it before the turn boundary, and the soft families ($A$, $C$, $E$) decay steadily once issues enter the persistent list and structural choices stabilise;
(ii) under \textsc{All-probes-every-turn}, $B_{\text{visual}}$ \emph{grows} monotonically --- with no scheduling, the per-turn critic raises new style findings on every slide every turn and never declares one done;
(iii) the rewrite edits those findings trigger then break layout, lifting $B_{\text{geom}}$ in step with $B_{\text{visual}}$ (annotated arrow in Fig.~\ref{fig:family-dynamics}, right);
this propagation of style-motivated edits into layout regressions, not any single hard family failing, drives the open-issue drift in Fig.~\ref{fig:dynamics}(a).
The factual family $D$ stays small in absolute count but has the highest \emph{reopen} rate per issue under \textsc{All-probes-every-turn}, because edits to neighbouring text break previously verified claims;
this is where step-level render feedback matters most beyond layout.
Per-family event-flow plots, reopen matrices, and the same decomposition for $N{\in}\{4,8\}$ are available in the released codebase.

\section{Diagnostic Mini-Ablations}
\label{app:diagnostics}

Tab.~\ref{tab:ablation-diag} reports two single-row diagnostics that the four-block ablation in Tab.~\ref{tab:ablation-quality} folds together: (i) splitting the per-turn NL-critique row by judge source (Self-Refine vs.\ Reflexion) and (ii) replacing the rendered-image input to the turn-level critic with the deck source code only.
GPT-5.4 $\times$ 100 tasks $\times$ 3 seeds, on the same scaffolding as Tab.~\ref{tab:ablation-quality}.

\begin{table}[h]
\centering
\small
\setlength{\tabcolsep}{4pt}
\caption{Single-row diagnostics for the per-turn NL-critique row of Tab.~\ref{tab:ablation-quality} (rows 1--2) and the turn-level critic input modality (row 3). Numbers are means over three seeds on the same $100$-task split as Tab.~\ref{tab:ablation-quality}.}
\label{tab:ablation-diag}
\begin{tabular}{l c c c c}
\toprule
Variant & Fid.\ $\uparrow$ & SCR $\uparrow$ & Aes.\ $\uparrow$ & Des.\ $\uparrow$ \\
\midrule
NL critique -- Self-Refine (self-judge)        & 76.5 & 67.8 & 3.15 & 60.5 \\
NL critique -- Reflexion (external judge)      & 77.1 & 68.6 & 3.21 & 60.5 \\
Critic text-only (no rendered image)           & 88.4 & 74.2 & 3.25 & 65.1 \\
\bottomrule
\end{tabular}
\end{table}

Rows~1--2 disaggregate the per-turn NL-critique row of Tab.~\ref{tab:ablation-quality} ($76.8/68.2/3.18/60.5$) into its two implementations: Self-Refine~\citep{selfrefine} uses the agent as its own critic, Reflexion~\citep{reflexion} uses an external judge.
The gap between them is below $1$ point on every module and within the observed seed-level variability, so the NL-critique modality, not its source, is the binding constraint.
Row~3 keeps the full \redeck{} loop but feeds the turn-level critic the deck source code only (no rendered image);
relative to the default \redeck{} configuration in Tab.~\ref{tab:ablation-quality} ($88.6/91.5/3.64/71.2$) this drops SCR by $17.3$ and Des.\ by $6.1$ while leaving Fid.\ approximately flat, validating the dual-space hypothesis (\S\ref{sec:method-3.3}) that text-only critique cannot localise spatial violations even when the action channel keeps step-level rendering intact.

\end{document}